\documentclass{article}

\PassOptionsToPackage{numbers, sort&compress}{natbib}

\usepackage[preprint]{neurips_2025}
\makeatletter
\renewcommand{\@noticestring}{Preprint. An earlier version was presented as a poster at the NASA 5th Eddy Cross-Disciplinary Symposium, Boulder, Colorado, May 2026.}
\renewcommand{\acksection}{\section*{Acknowledgments}}
\makeatother

\usepackage[utf8]{inputenc}
\usepackage[T1]{fontenc}
\usepackage{hyperref}
\usepackage{url}
\usepackage{booktabs}
\usepackage{amsfonts}
\usepackage{amsmath}
\usepackage{nicefrac}
\usepackage{microtype}
\usepackage{xcolor}
\usepackage{graphicx}
\usepackage{subcaption}
\usepackage{orcidlink}
\usepackage{float}
\usepackage{placeins}
\title{PISCES: Physics-Informed Solar-wind Convolutional autoEncoder for Space-weather Anomaly Detection and Early Warning}

\author{%
 Kevin Lee\orcidlink{0009-0004-0388-9260}$^{1,2,*}$\footnotemark[2] \quad Alison J. March\orcidlink{0009-0000-3643-8061}$^{3,*}$ \\
 [4pt] $^1$NASA Jet Propulsion Laboratory, La Ca\~{n}ada Flintridge, CA, USA \\
 $^2$Department of Mechanical and Aerospace Engineering, UCLA, Los Angeles, CA, USA \\
 $^3$College of Engineering \& Applied Science, University of Colorado, Boulder, CO, USA \\
}

\begin{document}

\begingroup
  \hypersetup{hidelinks, pdfborder={0 0 0}}
  \renewcommand{\thefootnote}{\fnsymbol{footnote}}
  \maketitle
  \footnotetext[1]{Equal contribution.}
  \footnotetext[2]{Corresponding author. Email: \texttt{kevinlee69720@g.ucla.edu}}
\endgroup

\begin{abstract}
Space weather early warning depends on detecting solar wind transients in in-situ measurements at the first Sun-Earth Lagrange point (L1), before they reach Earth. Fixed thresholds can miss combined magnetic and plasma structure, and many learning methods provide a single anomaly score. We present the Physics-Informed Solar-wind Convolutional autoEncoder for Space-weather (PISCES)%
\begingroup
  \renewcommand{\thefootnote}{\fnsymbol{footnote}}%
  \footnote[3]{Available at: \url{https://github.com/magnaprog/PISCES}. Additional files available upon request.}%
\endgroup,
a convolutional autoencoder trained without catalog labels on OMNI solar wind measurements under physics constraints. Its loss includes magnetic field consistency, an empirical relation between temperature and velocity, the Parker spiral angle, and penalties on changes between consecutive one-minute samples in derived quantities calculated from the reconstruction. At inference, PISCES separates the anomaly score into magnetic and plasma reconstruction errors, physics relations, and residual corrections, and reports the magnitude of each contribution. Attenuation of the skip connections, selected on validation data, improves average precision for the trained models, while the untrained scores remain nearly the same. The trained models also give a more consistent ordering of these physical contributions. After smoothing with a trailing median, the alarms can precede independently observed sudden commencements, including positive sudden impulses.
\end{abstract}


\section{Introduction}
\label{sec:intro}

When a coronal mass ejection reaches the Sun-Earth L1 Lagrange point, operators usually have around 40 minutes to an hour to assess the threat before the solar wind disturbance reaches the outer magnetosphere \cite{pulkkinen2007space}. In this narrow window, the decision to secure power grids, spacecraft electronics, and navigation systems depends on accurately distinguishing space weather transients from ordinary solar wind conditions \cite{pulkkinen2007space, hapgood2011towards}. Interplanetary coronal mass ejections (ICMEs), corotating interaction regions (CIRs), and interplanetary shocks leave distinct signatures in the magnetic field and plasma parameters at L1. Reliable real-time detection of each remains unresolved.

\begin{figure}[H]
  \centering
  \includegraphics[width=0.72\textwidth]{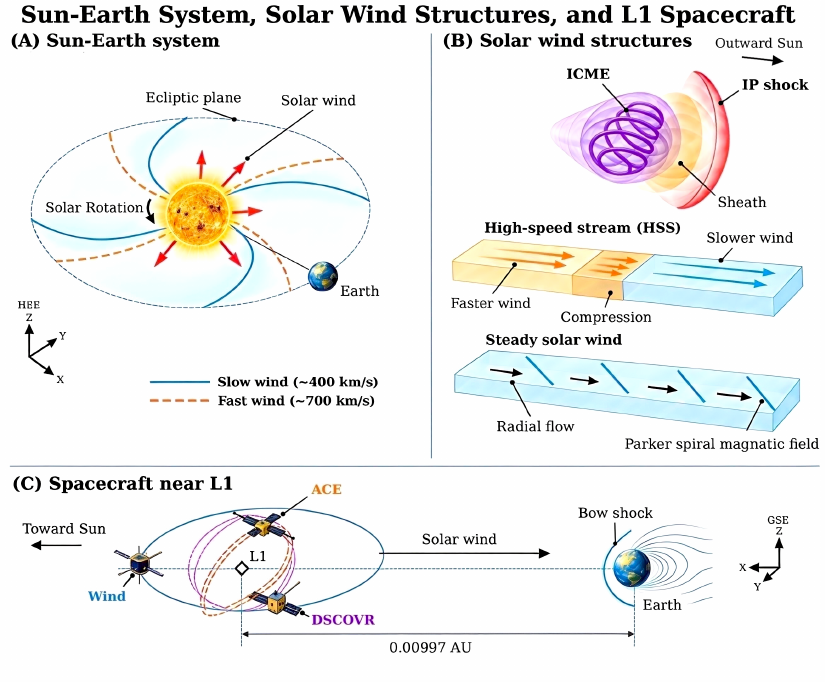}
\caption{Illustrative plots of Parker spiral geometry and solar wind structures observed near L1. \textbf{(A)} As the solar wind flows outward, solar rotation winds the interplanetary magnetic field into a spiral. The slow wind spiral is wound more tightly than the fast wind spiral. \textbf{(B)} The ICME carries magnetic structure outward and drives a shock. Compressed sheath plasma lies between the ICME and the shock. A high-speed stream overtakes slower wind and compresses the plasma ahead of it. In steady wind, the plasma flows radially outward, and the magnetic field lies at an angle to the flow. \textbf{(C)} ACE, Wind, and DSCOVR sample the solar wind on the sunward side of Earth, and the solar wind continues toward the bow shock. OMNI shifts the upstream observation times to the estimated arrival time at the nose of Earth's bow shock.}
  \label{fig:sun_earth}
\end{figure}

Real-time alerting at L1 still relies heavily on parameter thresholds. A southward $B_z$ below $-10$~nT for more than three hours is the classic criterion based on amplitude and duration for an intense geomagnetic storm \cite{gonzalez1994geomagnetic}. Systematic threshold studies of interplanetary shocks have found compression ratios of about 2.0 in density and about 1.9 in magnetic field \cite{cash2014characterizing}. These ratios characterize individual jumps but cannot represent how magnetic and plasma structures combine within one transient. Current supervised machine learning methods include long short-term memory (LSTM) networks for predicting geomagnetic indices \cite{gruet2018multiple}, Gaussian processes for classifying solar wind \cite{camporeale2017classification}, and deep segmentation networks for ICME boundary detection \cite{rudisser2022automatic, rudisser2026arcane}. These methods require labeled data. Existing catalogs are incomplete: Richardson \& Cane estimate that potential events with marginal signatures, which might or might not be included, constitute about 10\% of the events in their catalog \cite{richardson2010near}. We have not investigated uncatalogued events in this work. Unsupervised approaches can flag anomalies without catalog labels, including transient types missing from the catalogs. No prior work has yet used physics-informed loss regularization within an unsupervised autoencoder for in-situ solar wind data (Section~\ref{sec:related}).

We propose Physics-Informed Solar-wind Convolutional autoEncoder for Space-weather (PISCES), a physics-informed convolutional autoencoder using unlabeled solar wind data (Section~\ref{sec:data}). Its training loss combines an identity for total magnetic field magnitude, four penalties on changes between consecutive one-minute samples of reconstructed quantities, an empirical temperature and velocity relationship, and a geometric condition based on the Parker spiral (Section~\ref{sec:physics_loss}). PISCES reports constituent physical signatures and their magnitudes for each flagged window (Section~\ref{sec:scoring}).

\section{Background}
\label{sec:related}

Supervised methods have been used to detect solar wind events. R{\"u}disser et al.~\cite{rudisser2022automatic} built a ResUNet++ segmentation network trained on labeled ICME boundaries in in-situ observations from Wind and the Solar Terrestrial Relations Observatory (STEREO) spacecraft STEREO-A and STEREO-B. Their ARCANE system \cite{rudisser2026arcane} places that network in a streaming framework for early detection. A contemporary residual U-Net \cite{chen2022runet} addressed ICME detection with a similar architecture. Nguyen et al.~\cite{nguyen2019automatic} applied a convolutional neural network to ICME detection at two working points. The point with high recall detects 197 of the 232 ICMEs in the test period, with $84\%$ recall and $51\%$ precision. The point with high precision reaches $84\%$ precision with 25 false positives among 158 predictions, 62\% recall, and 145 of the 232 ICMEs detected. A later encoder-decoder approach jointly detects ICMEs and stream interaction regions in OMNI data \cite{nguyen2025simultaneous}. Camporeale et al.~\cite{camporeale2017classification} showed Gaussian process classification of solar wind types from in-situ plasma and field measurements together with indices of solar activity. These methods rely on labeled data. Their reliability for event types absent from the labels is unknown, and they lack a physical decomposition of \emph{why} a window is flagged.

Unsupervised methods avoid catalog labels. Anomaly detection based on reconstruction with autoencoders \cite{chen2017outlier} and variational autoencoders \cite{an2015variational} treats segments that reconstruct poorly as anomalous. Amaya et al.~\cite{amaya2020visualizing} combined autoencoders with dynamic self-organizing maps (DSOM) to classify solar wind regimes without labels and to represent the wind's multivariate structure. Physics-informed neural networks (PINNs) embed residuals of governing differential equations in the loss. Raissi et al.~\cite{raissi2019physics} showed that these constraints can regularize learning from small data sets, with applications in fluid dynamics \cite{cai2021physics} and laboratory plasma modeling \cite{mathews2021uncovering}. Physics constraints also appear in supervised solar wind forecasting losses \cite{johnson2023physics, johnson2024combining}. Physics-informed autoencoders with embedded physical laws have been used for unsupervised anomaly detection on power distribution grids \cite{zideh2024physics}. We know of no comparable autoencoder that applies a physics loss to in-situ solar wind data. Balancing reconstruction and physics losses is a multitask learning problem. Kendall et al.~\cite{kendall2018multi} proposed uncertainty weights for each task, and Wang et al.~\cite{wang2021understanding} studied gradient dynamics in physics-informed loss functions. We tried Kendall's uncertainty weighting for PISCES. These change penalties are zero when their associated derived quantities are constant across a reconstructed window (Section~\ref{sec:ablation}), and fixed weights were more stable.

PISCES combines an autoencoder for unsupervised anomaly detection, solar wind physics constraints in the loss, and an anomaly score decomposed into physically interpretable components.

\section{Method}
\label{sec:method}

\subsection{Data and Preprocessing}
\label{sec:data}

We use the Level-3 High Resolution OMNI (HRO) one-minute data set \cite{papitashvili2020omni}, in which upstream measurement timestamps are time-shifted to their estimated arrival time at Earth's bow shock nose. The Parker cone-angle relation defined in Section~\ref{sec:physics_loss} provides the geometric condition used in the loss. We use seven parameters. The magnetic components are $B_x$, $B_y$, and $B_z$ in Geocentric Solar Ecliptic (GSE) coordinates, and $B_t$, the magnitude calculated as $\sqrt{B_x^2 + B_y^2 + B_z^2}$. The plasma parameters are proton number density $n_p$, bulk speed $V$, and proton temperature $T_p$.

We split the data chronologically to avoid leakage. Training spans 2005 to 2015, from the decline of solar cycle 23 to the peak of cycle 24. Validation spans 2016 to 2017, and testing spans 2018 to 2024. Appendix~\ref{app:data} summarizes the split. Random splitting would put nearly duplicate windows on both sides of the boundary. It would also split corotating structures that recur about every 27 days of solar rotation and have predictive value at that lag \cite{schrijver2010evolving,owens2013persistence}. This would inflate results \cite{camporeale2019challenge}. A remaining source of dependence between the sets is a corotating structure crossing a split boundary, an edge effect affecting under one percent of the data. The test period covers the decline of solar cycle 24 and the rise of cycle 25.

The variables with positive values, $n_p$, $V$, and $T_p$, are transformed with a logarithm before standardization. After reconstruction, the inverse transform guarantees positivity without an additional realizability penalty. The magnetic field components, which may be negative in physical units, are standardized. Standardization statistics are computed only on the training set. The standardized time series are divided into 60-minute windows, advanced 15 minutes at a time in training and one minute at a time for evaluation and real-time scoring. For gaps of up to 15 minutes, we forward fill at most two samples, then interpolate any remaining missing values in time. Gaps longer than 15 minutes are restored to not-a-number (NaN) values, and windows containing them are dropped.

\subsection{Model Architecture}
\label{sec:architecture}

\begin{figure}[H]
  \centering
  \includegraphics[width=\textwidth]{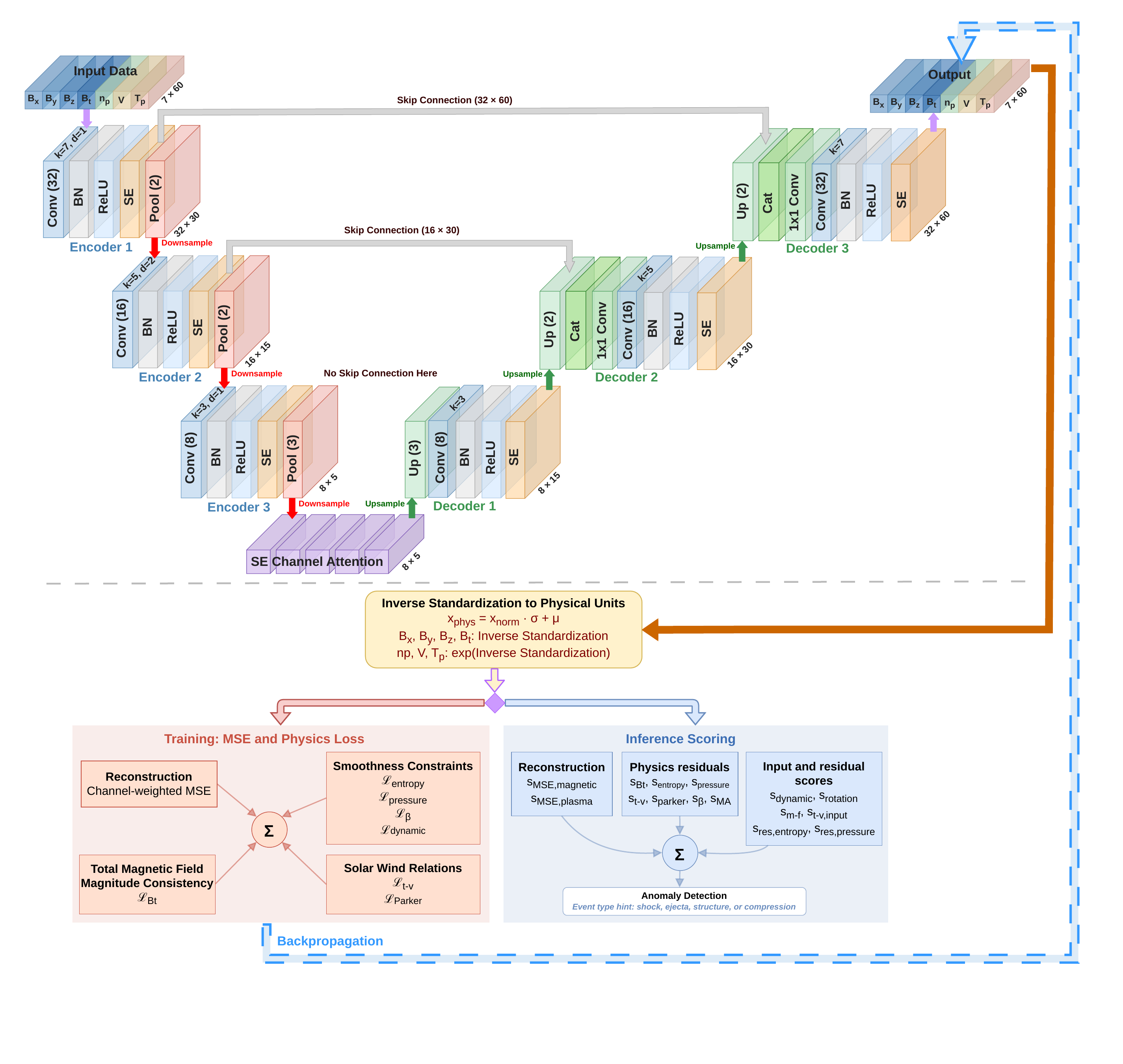}
\caption{PISCES architecture. \textbf{Top:} The encoder compresses the input window through three convolutional blocks to the bottleneck. The decoder reverses this structure. Skip connections enter through convolutions at the upper two levels, and the deepest level has no skip connection. Skip dropout occurs only during training. At inference, the model uses full skips by default. Table~\ref{tab:skip_interventions} reports attenuated inference. \textbf{Bottom:} During \emph{training} (left), reconstruction loss weighted by channel is combined with seven physical constraints. During \emph{inference} (right), the input and reconstruction produce 15 sub-scores, combined into an anomaly score with constant weights. Diagram labels: Conv, convolution; BN, batch normalization; ReLU, rectified linear unit; SE, squeeze-and-excitation; Pool, max pooling; Up, upsampling; Cat, concatenation; MSE, mean squared error.}
  \label{fig:architecture}
\end{figure}

PISCES is a one-dimensional convolutional autoencoder with U-Net skip connections \cite{ronneberger2015unet}, implemented in PyTorch \cite{paszke2019pytorch}. It processes seven input channels over 60-minute windows. Figure~\ref{fig:architecture} shows the architecture.

\paragraph{Encoder:}
Three one-dimensional convolutional blocks, each followed by batch normalization, a rectified linear unit (ReLU), and max pooling, reduce the temporal dimension from 60 to 30, then 15, and finally 5 steps with pooling factors of 2, 2, and 3, respectively. The blocks use 32, 16, and 8 filters with kernel sizes of 7, 5, and 3, dilations 1, 2, and 1, and same padding. A squeeze-and-excitation (SE) \cite{hu2018squeeze} mechanism uses global average pooling to reweight channels in each encoder block and in the bottleneck. The bottleneck representation has shape $(8, 5) = 40$ values for a compression ratio of $10.5{:}1$ from the $7 \times 60 = 420$ input values.

\paragraph{Decoder:}
Three upsampling blocks with factors 3, 2, and 2 reverse the encoder, with kernel sizes 3, 5, and 7 and squeeze-and-excitation attention at each level. Skip connections from the encoder enter the two upper levels through $1{\times}1$ convolutions, while the deepest decoder level has no skip connection. A final linear convolution returns the network to 7 channels. PISCES-Opto has 12,555 parameters, with decoder kernels reversed relative to the encoder's. PISCES-Mono is a comparison variant with decoder kernel size 3 throughout, no decoder attention, and 9,475 parameters.

\paragraph{Skip Connection Dropout:}
During training, standard dropout with $p=0.8$ is applied to the feature map for each skip connection. This randomly zeros 80\% of skip feature values and rescales the retained values by $1/(1-p)$. It discourages reliance on individual skip features. The bottleneck has a geometric compression ratio of $10.5{:}1$. At inference, dropout is off and the full skip connections carry detail from earlier encoder levels, as in a conventional U-Net. Section~\ref{sec:skip_interventions} shows that attenuation selected on validation data improves performance, while a smaller architecture retrained without skip or fusion parameters performs much worse. Within the tested setups, training with skip dropout and attenuating the skip contribution at inference yields higher average precision than training without skip connections.

\subsection{Physics-Informed Loss}
\label{sec:physics_loss}

All physics losses are defined over denormalized reconstructions through a differentiable inverse normalization layer embedded in the model. This layer reverses standardization and exponentiates channels transformed with a logarithm. It returns values in the original physical units (nT, cm$^{-3}$, km/s, K). PINNs embed governing differential equations in their loss. PISCES uses one identity equation for $B_t$ consistency, one empirical relationship between $T$ and $V$, one geometric condition based on the Parker spiral, and four penalties on changes between consecutive one-minute samples in the logarithms of derived quantities: entropy, pressure, beta, and dynamic pressure. These terms regularize the reconstruction; they do not assert that solar wind transients vary gradually.

\paragraph{Constraint 1: Magnetic Field Vector Consistency:}
The total magnetic field magnitude is defined by $B_t^2 = B_x^2 + B_y^2 + B_z^2$. The residual is penalized relative to $B_t^2$, with a $\max(B_t^2, 1\,\text{nT}^2)$ floor that prevents gradient singularities as $B \to 0$. Above the floor, the normalized residual is invariant to a uniform rescaling of the field:
\begin{equation}
\mathcal{L}_{\text{Bt}} = \frac{1}{NT}\sum_{i,t} \left(\frac{B_{t,\text{recon}}^2 - B_{x,\text{recon}}^2 - B_{y,\text{recon}}^2 - B_{z,\text{recon}}^2} {\max(B_{t,\text{recon}}^2,\; 1\,\text{nT}^2)}\right)^2
  \label{eq:loss_bt}
\end{equation}
The magnetic field magnitude $B$ of the OMNI vector is used instead of the scalar average of the magnitudes.

\paragraph{Constraint 2: Change Penalty for Proton Specific Entropy:}
For $\gamma = 5/3$, the proton specific entropy $S_p = T_p / n_p^{2/3}$ is among the parameters Xu \& Borovsky \cite{xu2015new} used to classify solar wind plasma types. The loss is the mean squared change in $\log S_p$ between consecutive one-minute samples:
\begin{equation}
\mathcal{L}_{\text{ent}} = \frac{1}{N(T{-}1)}\sum_{i,t} \left(\log S_{p,\text{recon}}^{(t+1)} - \log S_{p,\text{recon}}^{(t)}\right)^2
  \label{eq:loss_entropy}
\end{equation}
The term acts on reconstructed windows. It is an empirical regularizer, not a conservation law. Xu \& Borovsky \cite{xu2015new} show differences in specific entropy among their plasma types, with overlapping distributions.

\paragraph{Constraint 3: Change Penalty for Total Pressure:}
For total pressure $P_{\text{tot}} = n_p k_B T_p + B_t^2/(2\mu_0)$, the loss is the mean squared change in $\log P_{\text{tot}}$ between consecutive one-minute samples:
\begin{equation}
\mathcal{L}_{\text{pres}} = \frac{1}{N(T{-}1)}\sum_{i,t} \left(\log P_{\text{tot,recon}}^{(t+1)} - \log P_{\text{tot,recon}}^{(t)}\right)^2
  \label{eq:loss_pressure}
\end{equation}
This empirical regularizer does not treat total pressure as constant within a plasma type. The classifications in \cite{xu2015new, camporeale2017classification} report systematic thermodynamic differences among plasma types, while total pressure is not constant within a type. The sharp rise at shocks is well known \cite{cash2014characterizing}. The loss uses $k_B = 1.3807 \times 10^{-8}$~nPa$\cdot$cm$^3$/K and $1/(2\mu_0) = 3.9789 \times 10^{-4}$~nPa/nT$^2$ (Appendix~\ref{app:constants}). The thermal term uses proton temperature only because OMNI does not supply electron temperature. It underestimates the combined proton and electron thermal pressure by about a factor of 2 if their densities and temperatures are comparable. This penalty on changes in total pressure remains an empirical choice.

\paragraph{Constraint 4: Temperature and Velocity Relationship:}
Proton temperature has an empirical relationship with bulk speed in ambient solar wind \cite{lopez1987solar}. We use the linear screening fit from Elliott et al.\ \cite{elliott2012temporal}: $T_{\text{exp}}(\text{K}) = 486.5\,V(\text{km/s}) - 124{,}760$. The fit normalizes temperatures to a common heliocentric distance, screens periods with low beta, and regresses normalized temperature on speed. The $T$ and $V$ relationship changes from year to year, and fast wind in coronal holes follows a different line from slow wind. The screening fit is an approximation. $T_{\text{exp}}$ is a screening reference, not a fitted predictive relationship, and remains fixed in the loss and score calculations. ICMEs can have abnormally low temperatures for their speed. Richardson \& Cane \cite{richardson1995regions} use $T_p/T_{\text{exp}} \le 0.5$, with their own Lopez-based expected-temperature relation, to identify low-temperature plasma intervals. Here, we apply the same numerical cutoff to the Elliott et al. screening fit. Plasma in CIR compression regions can be hotter than this reference. The loss penalizes the deviation in logarithmic space:
\begin{equation}
\mathcal{L}_{\text{tv}} = \frac{1}{NT}\sum_{i,t} \left(\log T_{p,\text{recon}}^{(t)} - \log T_{\text{exp}}(V_{\text{recon}}^{(t)})\right)^2
  \label{eq:loss_tv}
\end{equation}

\begin{figure}[H]
  \centering
  \includegraphics[width=0.85\textwidth]{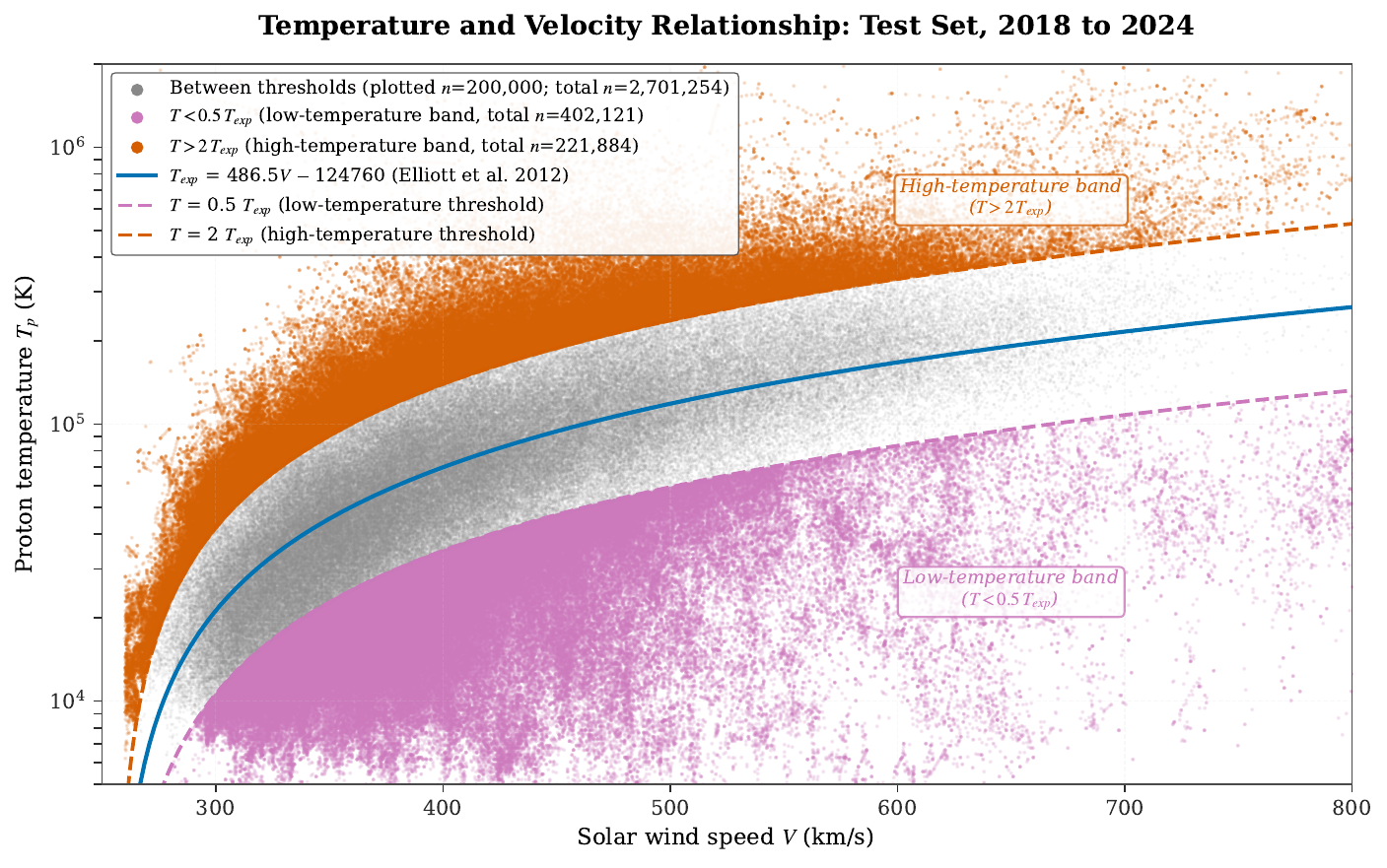}
\caption{Proton temperature versus solar wind speed for one-minute samples in the 2018 to 2024 test period after preprocessing. The plot retains finite ($V$, $T_p$) samples with $V > 260$~km/s and $T_p > 10^3$~K, about 3.3 million samples. The solid blue line is $T_{\text{exp}} = 486.5\,V - 124{,}760$\,K \cite{elliott2012temporal}. Dashed lines mark the low-temperature threshold $0.5 T_{\text{exp}}$ \cite{richardson1995regions}, shown in purple, and the high-temperature threshold $2\,T_{\text{exp}}$, shown in orange. Gray, purple, and orange dots account for 81\%, 12\%, and 7\% of the samples. A random subsample of 200,000 points between the two temperature thresholds makes the visual density readable. All cold and hot points appear, and percentages use the full filtered data.}
  \label{fig:tv_scatter}
\end{figure}

The loss encourages reconstructed plasma to follow the $T$ and $V$ correlation. This can increase reconstruction error for input plasma that departs from that relationship, as in ICMEs or compressions. Figure~\ref{fig:tv_scatter} shows filtered data from 2018 to 2024 with finite $V$ and $T_p$, $V>260$~km/s, and $T_p>10^3$~K. Twelve percent of samples lie below the $0.5\,T_{\text{exp}}$ threshold. The dashed bands lie a factor of two above and below $T_{\text{exp}}$, symmetric in $\log T_p$, where Eq.~\ref{eq:loss_tv} calculates the deviation. The lower band marks low-temperature plasma, which can occur in ICME ejecta. It does not identify a cataloged event.

\paragraph{Constraint 5: Parker Spiral Cone Angle:}
The Parker spiral \cite{parker1958dynamics} results from solar rotation acting on the interplanetary magnetic field. At 1~AU, the predicted Parker cone angle relative to the Sun-Earth line is $\psi = \arctan(\Omega_\odot r / V)$, where $\Omega_\odot r \approx 428.7$~km/s follows from the sidereal Carrington rotation period of 25.38 days. ICMEs have their own internal magnetic geometry and deviate systematically from the Parker spiral. The loss penalizes the discrepancy between the cone angle from the Sun-Earth line, $\arccos(|B_x|/B_t)$ in GSE coordinates, and the Parker cone angle in the ecliptic plane estimated from reconstructed speed. The cone angle comparison does not constrain field azimuth around the Sun-Earth line, so agreement with the predicted angle does not require an ecliptic field. Quiet wind can also depart from the predicted angle (Figure~\ref{fig:score_decomposition}). It is a soft constraint on deviation from the spiral:
\begin{equation}
\mathcal{L}_{\text{parker}} = \frac{1}{NT}\sum_{i,t} \left(\arccos\frac{|B_{x,\text{recon}}|}{B_{t,\text{recon}}} - \arctan\frac{\Omega_\odot r}{V_{\text{recon}}}\right)^2
  \label{eq:loss_parker}
\end{equation}

\paragraph{Constraint 6: Change Penalty for Plasma Beta:}
The loss is the mean squared change in the logarithm of proton plasma beta $\beta = P_{\text{th}} / P_B$ between consecutive one-minute samples. This empirical regularizer does not assume that beta is constant within an undisturbed stream. In quiet conditions at 1~AU, $\beta$ is approximately 0.5 to 2. Beta drops in ICMEs ($\beta \ll 1$) and varies rapidly in CIR compression regions. Xu \& Borovsky \cite{xu2015new} demonstrate that ejecta have lower $\beta$ than other plasma types. This difference is between types, and beta is not constant within a type. Low $\beta$ is one of the main characteristics of magnetic clouds \cite{burlaga1981magnetic, klein1982interplanetary}. Magnetic clouds constitute on average one quarter of near-Earth ICMEs from 1996 to 2002 \cite{cane2003interplanetary}. Their fraction swings sharply with the solar cycle, reaching 100\% during solar minimum and 15\% during solar maximum. The loss is:
\begin{equation}
\mathcal{L}_{\beta} = \frac{1}{N(T{-}1)}\sum_{i,t} \left(\log \beta_{\text{recon}}^{(t+1)} - \log \beta_{\text{recon}}^{(t)}\right)^2
  \label{eq:loss_beta}
\end{equation}

\paragraph{Constraint 7: Change Penalty for Dynamic Pressure:}
We use the dynamic pressure approximation $P_{\text{dyn}} = \frac{1}{2}m_p n_p V^2$. Space weather forecasts generally use flow pressure without the factor of $1/2$. The factor cancels in logarithmic differences, so we retain $\tfrac{1}{2}m_p n_p V^2$. The loss is the mean squared change in $\log P_{\text{dyn}}$ between consecutive one-minute samples. This empirical regularizer does not assume dynamic pressure is constant. Interplanetary shocks produce jumps in density and bulk speed \cite{cash2014characterizing}. The loss is:
\begin{equation}
\mathcal{L}_{\text{dyn}} = \frac{1}{N(T{-}1)}\sum_{i,t} \left(\log P_{\text{dyn,recon}}^{(t+1)} - \log P_{\text{dyn,recon}}^{(t)}\right)^2
  \label{eq:loss_dyn}
\end{equation}
using $P_{\text{dyn}}(\text{nPa}) = 8.363 \times 10^{-7} \times n_p(\text{cm}^{-3}) \times V^2(\text{km/s})^2$ (Appendix~\ref{app:constants}).

The total loss combines a mean squared error (MSE) reconstruction term weighted by channel with the physics term. The magnetic channel weights are 1.5, 1.5, 2.0, and 1.5 for $B_x$, $B_y$, $B_z$, and $B_t$. The plasma channel weights are 1.0, 1.2, and 1.0 for $n_p$, $V$, and $T_p$. The physics term carries a weight that ramps up over training and then stays fixed:
\begin{align}
\mathcal{L}_{\text{total}} &= \mathcal{L}_{\text{mse}} + w(\text{epoch})\,\mathcal{L}_{\text{phys}}
  \label{eq:loss_total} \\
\mathcal{L}_{\text{phys}} &= \lambda_{\text{Bt}}\,\mathcal{L}_{\text{Bt}} + \lambda_{\text{ent}}\,\mathcal{L}_{\text{ent}} + \lambda_{\text{pres}}\,\mathcal{L}_{\text{pres}} + \lambda_{\text{tv}}\,\mathcal{L}_{\text{tv}} + \lambda_{\text{parker}}\,\mathcal{L}_{\text{parker}} + \lambda_{\beta}\,\mathcal{L}_{\beta} + \lambda_{\text{dyn}}\,\mathcal{L}_{\text{dyn}}
  \label{eq:loss_phys}
\end{align}
where $\lambda_{\text{Bt}} = 5{\times}10^{-4}$, $\lambda_{\text{ent}} = 0.5$, $\lambda_{\text{pres}} = 0.3$, $\lambda_{\text{tv}} = 0.003$, $\lambda_{\text{parker}} = 3{\times}10^{-4}$, $\lambda_{\beta} = 0.03$, $\lambda_{\text{dyn}} = 0.05$.

The ramp factor $w(\text{epoch})$ is zero during the initial 10-epoch warmup with only MSE. It rises linearly from epoch 11 to 20 in ten equal increments from $0.1\,pw_{\max}$ to $pw_{\max}$, then stays fixed. The gradual ramp avoids instability from the sudden addition of physics gradients. The effective weights on physics losses satisfy $w(\text{epoch})\,\lambda_i \le pw_{\max}\,\lambda_i$.

The PISCES-Opto checkpoint uses the same seven loss terms with $pw_{\max}=0.30$ and reaches its lowest validation MSE of 0.02196 at epoch 95. Because this checkpoint was selected after inspecting corrected test metrics, it is a released checkpoint, not blind test evidence. For numerical stability, positive values are clamped to at least a small $\varepsilon > 0$, bulk speed to at least 200\,km/s, and expected temperature to at least 1000\,K before logarithms are taken.

\subsection{Anomaly Scoring}
\label{sec:scoring}

At inference, each 60-minute window gives a score vector comprising fifteen sub-scores.

Two reconstruction error metrics group the MSE by channel: $s_{\text{mag}}$ is the reconstruction error of the magnetic channels ($B_x, B_y, B_z, B_t$) and $s_{\text{plasma}}$ is the reconstruction error of the plasma channels ($n_p, V, T_p$).

There are seven sub-scores corresponding to the seven physics constraints used in the training loss. Six are calculated from the reconstruction: the magnetic field consistency $s_{\text{Bt}} = \text{mean}\,|B_t^2 - B_x^2 - B_y^2 - B_z^2|$, the raw residual in nT$^2$ rather than the dimensionless normalized training loss of Eq.~\ref{eq:loss_bt}; the entropy jump $s_{\text{ent}} = \max|\Delta\log S_p|$; the total pressure jump $s_{\text{pres}} = \max|\Delta\log P_{\text{tot}}|$; the temperature and velocity relationship $s_{\text{tv}} = \max|\log T_p - \log T_{\text{exp}}(V)|$; the Parker spiral angle $s_{\text{parker}} = \text{mean}\,|\theta_{\text{cone}} - \theta_{\text{expected}}|$; and the plasma beta jump $s_{\beta} = \max|\Delta\log\beta|$. The seventh, $s_{\text{dyn}} = \max|\Delta\log P_{\text{dyn}}|$, measures dynamic pressure jumps on the raw input instead of the reconstruction.

Deviation of the Alfv\'{e}n Mach number from its reference value, $s_{M_A} = \text{mean}\,|\log M_A - \log 7|$, is calculated at inference without an analogue in the training loss since regularizing $M_A$ toward $7$ would suppress detection of sub-Alfv\'{e}nic ICME ejecta. The reference lies near the lower end of the ambient 1~AU range ($M_A \approx$ 6 to 11, median 9 to 10 \cite{klein2019solar,schrijver2009plasma}). The score measures the absolute logarithmic ratio to this reference.

Three further sub-scores are computed from the raw time series: the magnetic field rotation rate $s_{\text{rot}}$ (mean radians\,min$^{-1}$); the mass flux jump $s_{\text{mf}}=\max|\Delta\log(n_p V)|$; and the input temperature and velocity relationship $s_{\text{tv,in}} = \text{mean}\,|\log T_p - \log T_{\text{exp}}(V)|$.

The mass flux sub-score detects abrupt changes in the particle flux $n_p V$, which varies at interplanetary shocks, and the input $T$ and $V$ sub-score measures the input's deviation from the $T$ and $V$ relationship regardless of reconstruction quality. Two physics residual sub-scores measure the correction applied by the model: for entropy and total pressure, the residual is the window mean of $\big||\Delta_{\text{input}}| - |\Delta_{\text{recon}}|\big|$. A large residual means the physics signature of the reconstructed signal departs strongly from that of the input.

The composite score combines all fifteen sub-scores with fixed weights:
\begin{equation}
\begin{aligned}
s_{\text{comp}} = {}& s_{\text{mag}} + s_{\text{plasma}} + 0.005\,(s_{\text{Bt}} + s_{\text{ent}} + s_{\text{pres}} + s_{\text{dyn}} + s_{\text{res,ent}} + s_{\text{res,pres}}) \\ &+ 0.003\,(s_{\text{tv}} + s_{\beta} + s_{\text{rot}} + s_{\text{mf}} + s_{\text{tv,in}}) + 0.002\,(s_{\text{parker}} + s_{M_A})
\end{aligned}
  \label{eq:composite}
\end{equation}
These constant weights are used for the results and analysis below.

We tested weight optimization with differential evolution and rescaling each sub-score from its 50th to 99th validation percentile. It improved precision-recall area under the curve (PR-AUC) on validation data but degraded it on test data, consistent with overfitting about 172 validation events. For offline evaluation, a centered median filter over 15 minutes is applied to the composite before the 99th percentile threshold of the validation set is computed. Real-time use requires a trailing median filter and a threshold derived from scores produced by that filter.

\subsection{Hints of Event Types}

Hints of event types follow a fixed rule using the fractional contributions of $s_{\text{mag}}$ and $s_{\text{plasma}}$ to their sum, together with thresholds on the derived physics sub-scores. The conditions are evaluated in order, and the first match sets the hint. A dynamic pressure jump ($s_{\text{dyn}} > 0.3$) together with magnetic and plasma fractions both above 0.2 indicates an interplanetary shock. A low plasma $\beta$ ($< 0.5$) and a high field rotation rate ($s_{\text{rot}} > 0.1$~rad\,min$^{-1}$) indicate structures such as magnetic ejecta; at this threshold, $s_{\text{rot}}$ is sensitive to fast rotations at ejecta boundaries and in the sheath but not to slow rotation inside a flux rope. Magnetic fraction dominance ($> 0.4$) or a high $B_t$ consistency residual ($s_{\text{Bt}} > 2.0$) indicates magnetic structure. Plasma fraction dominance ($> 0.4$) or an entropy jump ($s_{\text{ent}} > 0.2$) without a dynamic pressure jump suggests a compression region.

\begin{figure}[p]
  \centering
  \includegraphics[width=\textwidth, height=0.70\textheight, keepaspectratio]{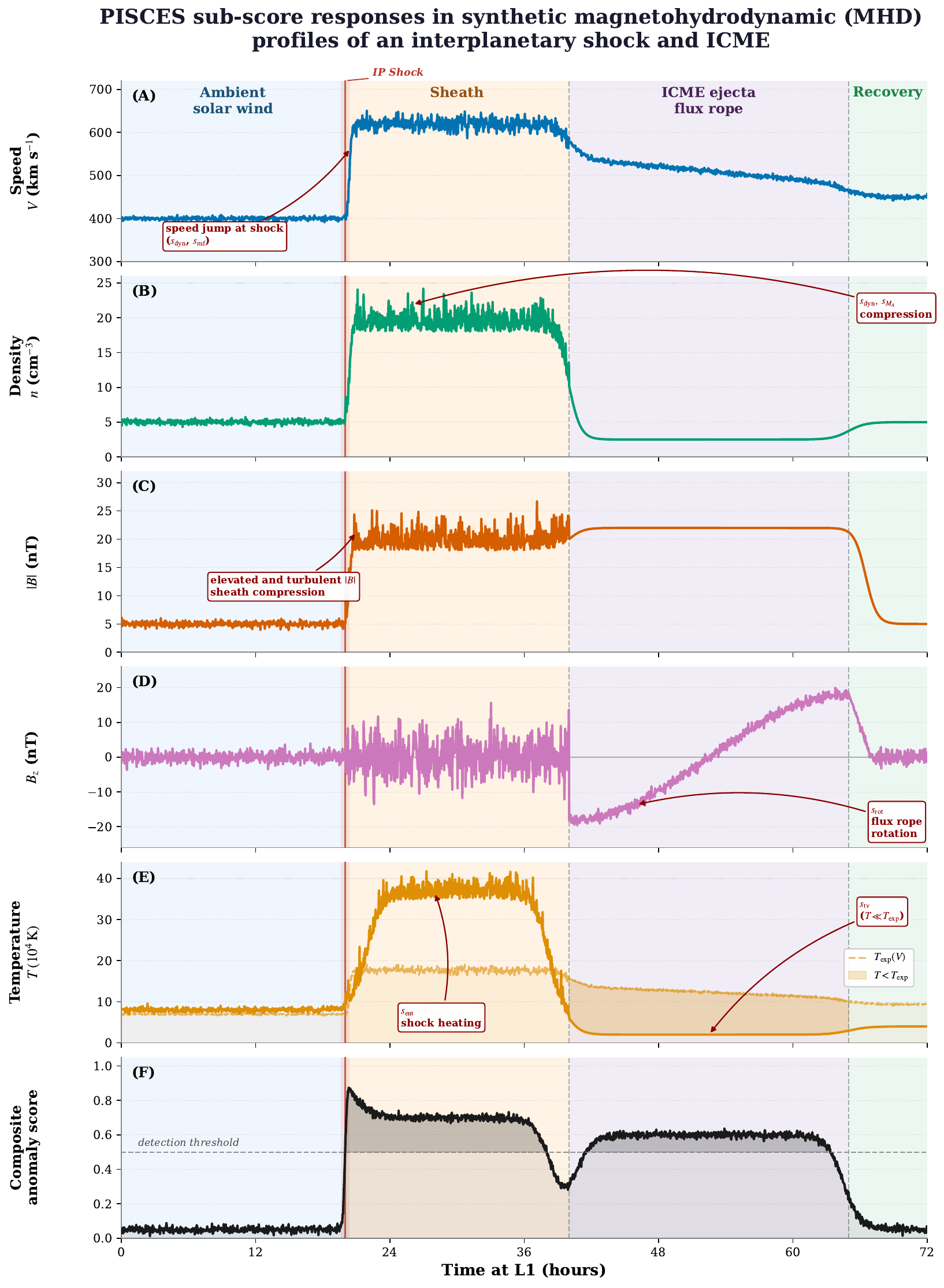}
\caption{PISCES sub-score responses for a synthetic magnetohydrodynamic (MHD) profile of an interplanetary shock and interplanetary coronal mass ejection (ICME) over 72 hours. Phase colors mark ambient solar wind, compressed sheath, ICME magnetic ejecta in a flux rope, and recovery. The red line marks interplanetary (IP) shock onset, and gray dashed lines enclose the ejecta. \textbf{(A)}~Solar wind speed $V$ jumps at the shock, driving dynamic pressure $s_{\mathrm{dyn}}$ and mass flux $s_{\mathrm{mf}}$ scores, then decays through the expanding ejecta. \textbf{(B)}~Proton density $n$ increases in the sheath and falls in the ejecta, driving $s_{\mathrm{dyn}}$ and Alfv\'{e}n Mach $s_{M_A}$. \textbf{(C)}~Total magnetic field magnitude $|B|$ is enhanced and turbulent in the sheath and remains high in the ejecta. The $s_{\text{Bt}}$ sub-score measures inconsistency among reconstructed field channels, not field enhancement itself. \textbf{(D)}~North-south magnetic field component $B_z$ rotates through the flux rope, with a sharp rotation at the leading edge and sheath turbulence that drive $s_{\mathrm{rot}}$. The $B_z$ component returns toward its ambient level over about two hours at the trailing edge. \textbf{(E)}~Proton temperature $T$ (solid) is anomalously low in the ejecta relative to $T_{\mathrm{exp}}(V)$ (dashed)~\cite{elliott2012temporal}. Orange shading marks $T < T_{\mathrm{exp}}$, a temperature deficit that sustains the temperature and velocity score $s_{\mathrm{tv}}$. The entropy-jump sub-score $s_{\mathrm{ent}}$ responds at the two interval boundaries. \textbf{(F)}~The composite anomaly score peaks at shock onset, remains highest in the sheath, and stays elevated through the smoother ejecta. Shading above the dashed detection threshold marks flagged intervals. Jump amplitudes are modeled on real events. The profile is illustrative and does not solve the Rankine-Hugoniot conditions, and the composite trace is not model output. Figure~\ref{fig:insitu_gannon} shows analogous observations from the May~2024 Gannon superstorm.}
  \label{fig:mhd_phase}
\end{figure}

Figure~\ref{fig:mhd_phase} illustrates how the 15 sub-scores correspond to the one-dimensional structure of an interplanetary shock and ICME at L1. Each subplot contains a synthetic profile modeled on a real event, with labels for the relevant sub-scores. The prescribed composite trace rises at shock arrival and remains elevated through the sheath and ejecta. It is illustrative and is not calculated from the sub-scores.

\section{Experiment Setup}
\label{sec:experiments}

\subsection{Training Specifications}
\begin{figure}[htbp]
  \centering
  \includegraphics[width=0.95\textwidth]{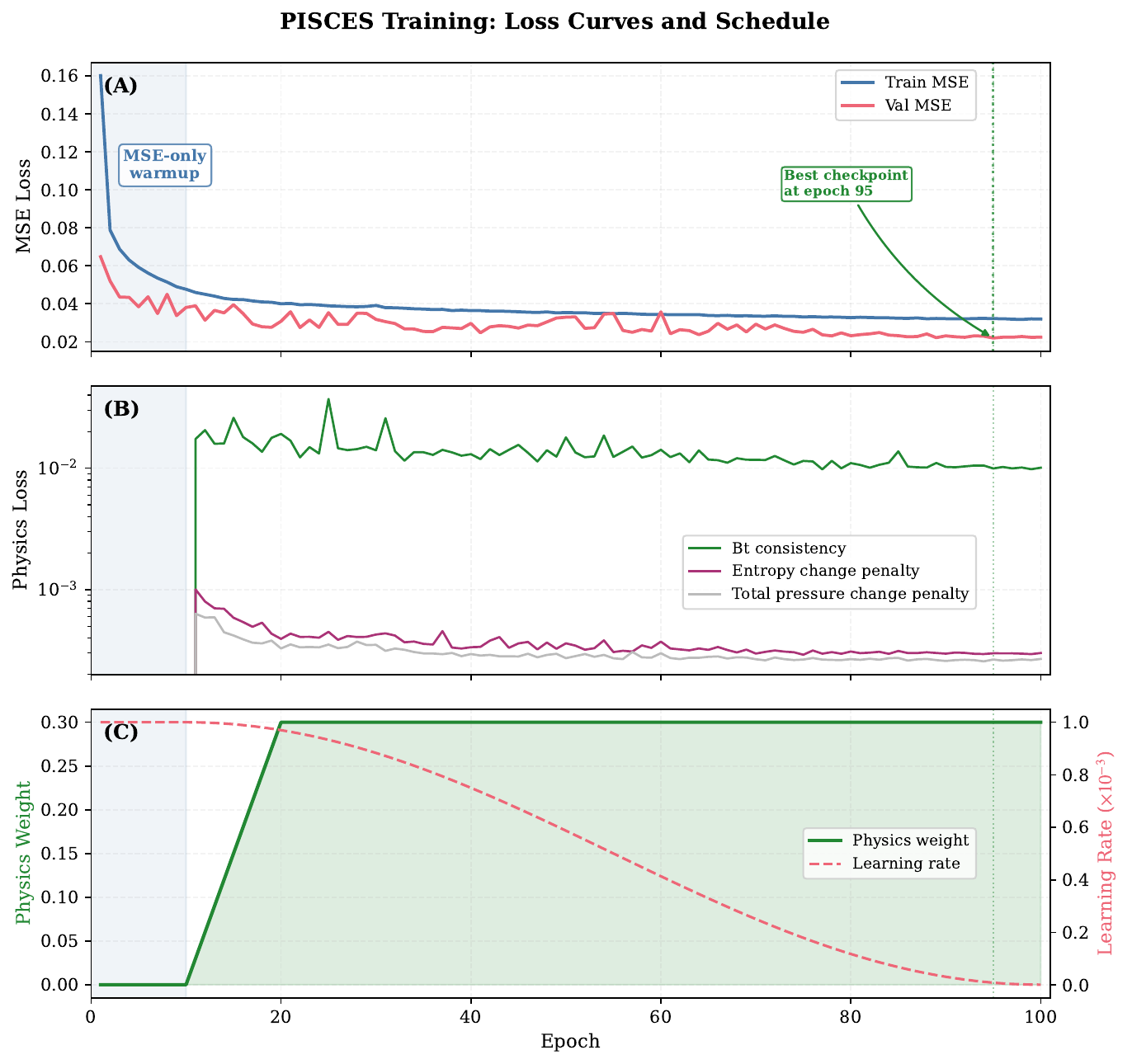}
\caption{Training loss curves and schedules for PISCES-Opto. The best validation mean squared error (MSE) is 0.02196 at epoch 95 of 100. Validation MSE is below training MSE in all but two epochs. Skip dropout is applied only during training. The shaded region shows the 10-epoch warmup using only MSE, and the dashed line marks the best checkpoint. \textbf{(A)}~Training and validation reconstruction MSE. \textbf{(B)}~The $B_t$ consistency loss and change penalties for entropy and total pressure, on a logarithmic scale after warmup. \textbf{(C)}~Ramp in physics weight to 0.30 and learning rate decay.}
  \label{fig:training_curves}
\end{figure}

We trained PISCES with the Adam optimizer at learning rate $10^{-3}$, batch size 256, gradient clipping at maximum norm 1.0, and cosine annealing from the end of warmup to a minimum learning rate of $10^{-6}$. Early stopping uses a patience of 25 epochs on validation MSE. When physics training starts, we reset the early stopping state and discard the best warmup checkpoint so the physics phase starts its own 25-epoch patience interval. Figure~\ref{fig:training_curves} shows the loss curves and schedule. We monitor validation MSE because the ramp in physics losses raises total loss even while reconstruction improves. After the 10-epoch warmup using only MSE, physics losses ramp from $0.1\,pw_{\max}$ to the configured $pw_{\max}$ in ten equal steps over epochs 11 to 20, as defined in Section~\ref{sec:physics_loss}. PISCES-Opto checkpoint uses $pw_{\max}=0.30$, skip connection dropout probability $0.8$, and weight decay $10^{-4}$. It reaches the best validation MSE of $0.02196$ at epoch 95. Results for PISCES-Mono and model ablations appear in Appendix~\ref{app:hyperparams}, which lists all hyperparameters.

\subsection{Evaluation Methodology}
We compare detections with NASA's Space Weather Database Of Notifications, Knowledge, Information (DONKI) catalogs of high-speed streams and interplanetary shocks (IPS) and the Richardson \& Cane near-Earth ICME catalog \cite{cane2003interplanetary, richardson2010near}. A $\pm 3$-hour match window accommodates catalog timing uncertainty of about one to two hours while limiting the interval in which unrelated alarms can be matched. Richardson \& Cane group low-temperature measurements separated by less than 3 hours into one plasma interval \cite{richardson1995regions}.
Appendix~\ref{app:match_window} shows sensitivity to this choice. For ICMEs, we use the ``ICME Plasma/Field Start'' and ``ICME Plasma/Field End'' entries in the Richardson \& Cane catalog, which record the start and end of each ICME plasma and field signature at the near-Earth spacecraft \cite{richardson2010near}. IP shock and high-speed stream windows receive positive labels when they fall within $\pm 3$ hours of the catalog event time. ICME windows receive positive labels from $\text{start} - 3\,\text{h}$ through $\text{end} + 3\,\text{h}$. This is an interval label because ICMEs are prolonged events. The 102 catalog intervals in 2018 to 2024 have a median length of 26 hours. The detection threshold is the 99th percentile of composite anomaly scores in the 2016 to 2017 validation set. PR-AUC traces precision and recall over all thresholds, avoiding dependence on one chosen threshold.

Point-adjusted $F_1$ can make random anomaly scores appear competitive \cite{kim2022rigorous}, and incorrect benchmarks for time series anomaly detection can create the same illusion of progress \cite{wu2023flawed}. PR-AUC uses scores from overlapping windows and depends on which windows the matching tolerance labels positive. PISCES and all baselines use the same labeling convention (Appendix~\ref{app:match_window}).

\subsection{Metrics}
Precision-Recall Area Under the Curve (PR-AUC) is the primary metric because ICME interval labels yield a positive window frequency of about 11.9\% \cite{davis2006relationship, saito2015precision}. We compute it as average precision, the sum of precision times incremental recall over descending score thresholds. We also report window probability of detection (POD, or recall), window false alarm ratio (FAR), precision, Heidke skill score, event POD, alarm event FAR, and detection offset from catalog onset at L1. Positive offsets indicate detection before catalog onset, and negative offsets indicate detection after onset. Event POD is the fraction of catalog events matched by at least one detection. Alarm event FAR is the fraction of contiguous alarm episodes that do not overlap an event interval. These measures resemble range-based evaluation \cite{tatbul2018precision}, though their precision and recall definitions differ.

\section{Results}
\label{sec:results}

\subsection{Performance Comparison}
\label{sec:main_results}

We evaluate PISCES-Opto and PISCES-Mono alongside baseline methods. Table~\ref{tab:main_results} gives the main results on the test set. Later figures show case studies and score decompositions.

Table~\ref{tab:main_results} presents scores from the 2018 to 2024 test set. We evaluate against the DONKI catalogs for interplanetary shocks (IPS) and high-speed streams and the Richardson \& Cane near-Earth ICME catalog \cite{cane2003interplanetary, richardson2010near}. We compare PISCES with one-class support vector machines (OC-SVM)~\cite{scholkopf2001estimating}, a convolutional variational autoencoder (Conv-VAE)~\cite{an2015variational}, principal component analysis (PCA) reconstruction~\cite{pearson1901principal}, a long short-term memory autoencoder (LSTM-AE)~\cite{hochreiter1997long,malhotra2016lstm}, and a maximum jump rule. The Conv-VAE score is the mean squared reconstruction error obtained by decoding the latent mean.

\begin{table}[t]
\caption{Detection performance on the test set in the period from 2018 to 2024 with 2,969,144 windows evaluated one minute apart. Baseline labels denote convolutional variational autoencoder (Conv-VAE), one-class support vector machine (OC-SVM), principal component analysis (PCA), and long short-term memory autoencoder (LSTM-AE). The catalog contains 863 events: 328 interplanetary shocks, 433 high-speed streams, and 102 interplanetary coronal mass ejections (ICMEs); 848 overlap the scored coverage and form the denominator for event coverage. We timestamp each window at its end, apply a validation 99th percentile threshold, and use centered median smoothing over 15 minutes within each contiguous segment. Shocks and high-speed streams are labeled by point; ICMEs are labeled by interval. Probability of detection (POD), false alarm ratio (FAR), precision, and Heidke scores are evaluated by window; event POD is the fraction of catalog events eligible for coverage matched by at least one alarm; episode FAR is the false alarm ratio in contiguous alarm episodes. PISCES-Opto reports precision-recall area under the curve (PR-AUC) of $0.365$ for the run with the highest score among the eleven training runs. Those scores range from $0.257$ to $0.365$ and have a mean of $0.306 \pm 0.029$. A bootstrap over blocks defined by calendar month gives a 95\% confidence interval from $0.306$ to $0.419$ for this checkpoint; it was selected after corrected test metrics were inspected, so it is not a blind test result. The released checkpoint uses an earlier normalization statistics artifact than the other ten, two older checkpoint files omit several training hyperparameters, and the standard deviation is therefore not a clean estimate of seed variation alone. Baselines are retrained and evaluated under the same procedure, without additional tuning. The Conv-VAE, OC-SVM, and LSTM-AE rows are means across three runs with seeds 7, 13, and 42, and their PR-AUC cells also give the sample standard deviation. Each remaining baseline metric is the mean across runs, with each run evaluated at its validation 99th percentile threshold. PCA reconstruction reproduces every displayed metric across all three runs. These standard deviations describe variation across runs on this fixed test set, not uncertainty from temporal variation or catalog sampling. The row of window energy scores each window by the mean square of its standardized inputs with the same procedure. It uses seven means and standard deviations from the training set but fits no model, and its PR-AUC of $0.341$ exceeds six of the seven method rows above it.}
  \label{tab:main_results}
  \centering
  \vspace{6pt}
  \resizebox{\textwidth}{!}{%
  \begin{tabular}{lccccccc}
    \toprule
    Method & PR-AUC & POD$_w$ & FAR$_w$ & Precision$_w$ & Heidke$_w$ & Event POD & Episode FAR \\
    \midrule
    PISCES-Opto checkpoint & 0.365 & 0.090 & 0.252 & 0.748 & 0.139 & 0.229 & 0.417 \\
    Mean $\pm$ sd across the eleven runs & $0.306 \pm 0.029$ & $0.058 \pm 0.013$ & $0.373 \pm 0.066$ & $0.627 \pm 0.066$ & $0.088 \pm 0.021$ & $0.201 \pm 0.016$ & $0.458 \pm 0.037$ \\
    PISCES-Mono & 0.322 & 0.056 & 0.352 & 0.648 & 0.086 & 0.183 & 0.469 \\
    \midrule
    Conv-VAE & $0.228 \pm 0.049$ & 0.041 & 0.490 & 0.510 & 0.060 & 0.199 & 0.546 \\
    OC-SVM (PCA, 95\% variance) & $0.251 \pm 0.004$ & 0.078 & 0.342 & 0.658 & 0.117 & 0.171 & 0.481 \\
    PCA reconstruction & 0.207 & 0.035 & 0.502 & 0.498 & 0.050 & 0.236 & 0.522 \\
    LSTM-AE & $0.195 \pm 0.001$ & 0.035 & 0.516 & 0.484 & 0.050 & 0.200 & 0.547 \\
    Max-Jump & 0.176 & 0.026 & 0.590 & 0.410 & 0.036 & 0.163 & 0.570 \\
    \midrule
    Window energy without a fitted model & 0.341 & 0.064 & 0.297 & 0.703 & 0.100 & 0.101 & 0.378 \\
    \bottomrule
  \end{tabular}%
}
\end{table}

Rerunning Conv-VAE with three seeds gives PR-AUC from $0.199$ to $0.285$, LSTM-AE from $0.194$ to $0.196$, and OC-SVM from $0.246$ to $0.254$. PCA reconstruction remains at $0.207$. None of the three Conv-VAE reruns reaches the earlier committed result of $0.287$. The run initialized with seed 42 gives $0.285$. The bottom row uses no fitted model. It scores each window by the mean square of its standardized values under the same procedure, using seven training set means and standard deviations. Rows with checkpoints use the normalization associated with their results. Its PR-AUC of $0.341$ exceeds six of the seven method rows above it and the mean across eleven PISCES runs.

\begin{figure}[htbp]
  \centering
  \includegraphics[width=0.85\textwidth]{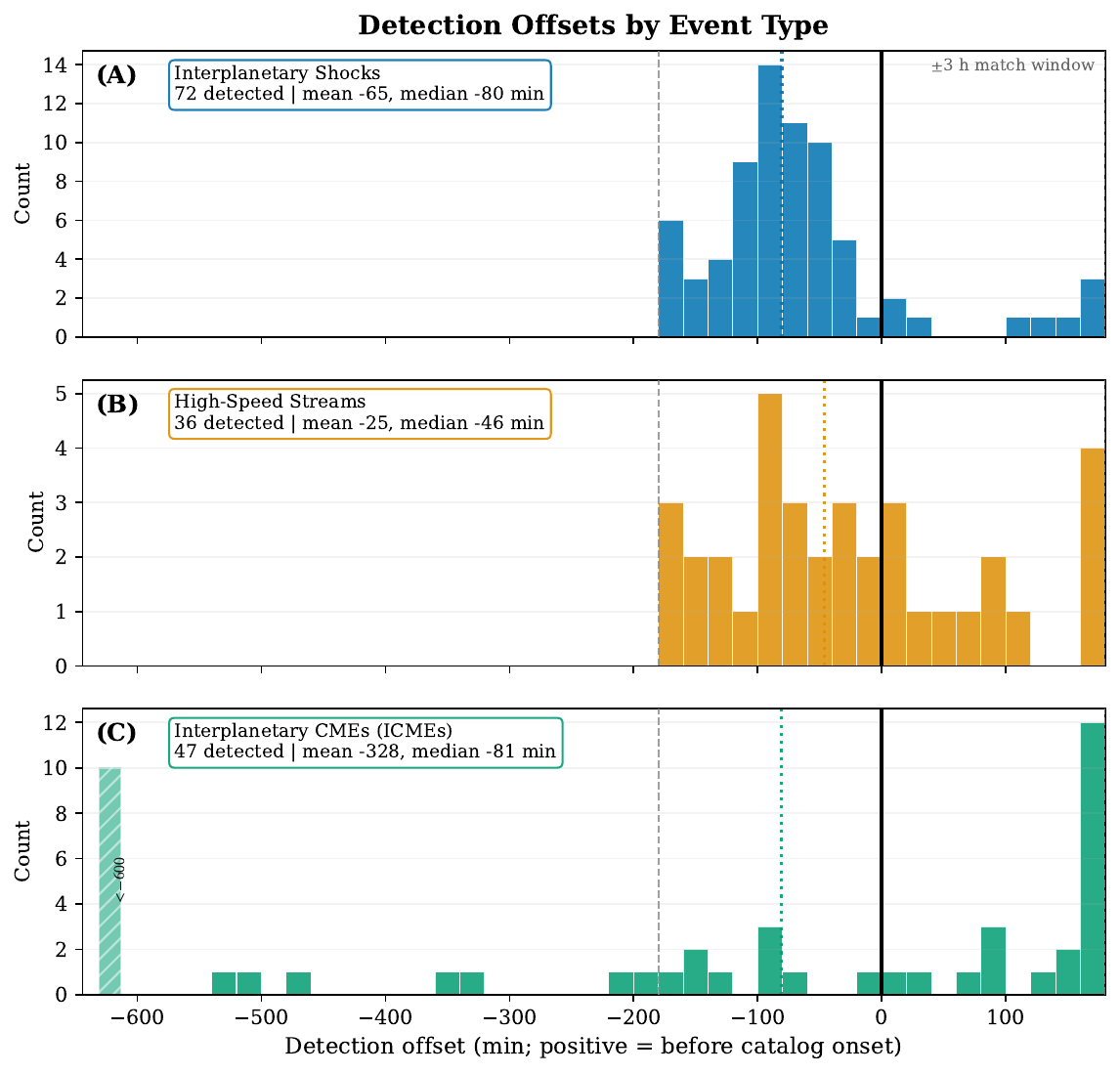}
\caption{Distribution of detection offsets from catalog event onset, in 20-minute bins, for interplanetary shocks, high-speed streams, and interplanetary coronal mass ejections (ICMEs) detected by PISCES-Mono. \textbf{(A)}~Interplanetary shocks. \textbf{(B)}~High-speed streams. \textbf{(C)}~ICMEs. The threshold is the 99th percentile value selected from 2016 to 2017 validation scores. Positive offsets indicate detection before catalog onset at L1. Negative offsets indicate detection after onset. Dotted lines mark the medians. Gray dashed lines show the $\pm3$~h match window for point events, so pileups near $\pm180$~min arise from that boundary. Hatched overflow bars contain 10 ICME detections more than 600 minutes from onset. A detection from 3 hours before the catalog plasma/field start through 3 hours after its end counts as an ICME hit, so a matched detection can occur several hours after the onset timestamp.}
  \label{fig:lead_times}
\end{figure}

At the 99th percentile threshold, PISCES-Opto flags 42,480 windows and matches 194 of the 848 catalog events eligible for event coverage. Mean and median offsets from the L1 catalog onset are $-133.5$ and $-66.5$ minutes, respectively; 29.4\% of matched detections occur before onset and 70.6\% occur after it. These numbers describe timing relative to catalog onset, not warning lead. Figure~\ref{fig:lead_times} shows the offset distribution by event type for PISCES-Mono. L1 measurements can precede geomagnetic impact at Earth, so detections during an event may still be operationally useful.

Interplanetary shocks are sudden jumps \cite{cash2014characterizing} that appear in OMNI data sampled once per minute as abrupt steps across channels. High-speed stream entries mark intervals of fast wind, often associated with stream interaction regions (SIRs). SIRs form where fast solar wind compresses preceding slow wind. Recurrent SIRs are termed corotating interaction regions (CIRs). The compressed region forms in the portion of the stream where speed increases, which typically lasts two to three days at 1~AU \cite{grandin2019properties}. At 1~AU, CIRs are usually unshocked because forward and reverse shocks form beyond Earth's orbit \cite{schrijver2010storms}, and the region changes gradually over a 60-minute window. This timescale mismatch helps explain the lower occurrence rate of high-speed stream detections; a $\pm3$-hour point label is only an approximation of a region that lasts for days.

ICMEs are associated with intense geomagnetic storms \cite{cane2003interplanetary, richardson2010near}. Their in-situ characteristics relate to several PISCES physics terms. Depressed proton temperature contributes to $\mathcal{L}_{\text{tv}}$ because ICMEs often violate the $T$ and $V$ relationship with $T_p < 0.5\,T_{\text{exp}}$ \cite{richardson1995regions}. Low plasma beta ($\beta \ll 1$) is characteristic of magnetic clouds, while changes in beta contribute to $\mathcal{L}_{\beta}$. Parker spiral deviation in magnetic clouds contributes to $\mathcal{L}_{\text{parker}}$. These terms are evaluated on reconstructed quantities, so observed ICME signatures do not by themselves establish their contribution to detection. ICME occurrence varies with phase of the solar cycle, from 4 to 27 events per year between 2018 and 2024, spanning the late cycle 24 minimum through the cycle 25 rise. Sharp shock boundaries can produce changes within a window more readily than gradual stream regions. A detection anywhere within an ICME passage lasting several hours counts as a hit.

Table~\ref{tab:per_event_type} shows the same ordering in event POD and receiver operating characteristic area under the curve (ROC-AUC). Event POD is 0.690 for ICMEs, 0.248 for interplanetary shocks, and 0.107 for high-speed streams. The positive labels amount to 26.2~h per retained ICME, compared with 4.9~h per shock and 5.0~h per stream. ROC-AUC follows the same ordering at 0.838, 0.655, and 0.598. This ordering is consistent with ICME intervals that are easier to cover. It does not establish the cause because interval labels confound the comparison. A random classifier scores 0.5 on ROC-AUC regardless of prevalence. ROC-AUC still depends on the windows labeled positive, and these rows do not isolate event physics, label duration, and overlap among event types.

\begin{table}[H]
\caption{PISCES-Opto performance by event type. As in Table~\ref{tab:main_results}, the table uses a 99th percentile threshold on the validation set, timestamps at window ends, and centered median smoothing over 15 minutes within contiguous score segments. The typed rows sum to 848 events eligible for coverage and 194 detected events. The subscript \textit{$w$} denotes window scores. Each typed row treats its event type as positive, so windows labeled only as another type are negative for that row. Precision-recall area under the curve (PR-AUC) and receiver operating characteristic area under the curve (ROC-AUC) for each type count correct detections of another type as false positives; this penalty is particularly large for weaker classes because their negative windows include ICME windows with high scores. There are 380,584 positive window assignments over a union of 352,278 windows; overlap accounts for the 28,306 excess assignments (7.4\% of typed positive assignments). The effective label coverage is 26.2~h per ICME, 4.9~h per shock, and 5.0~h per stream. These effective exposures are not physical event durations, so event probability of detection (POD) is not directly comparable between point and interval labels. PR-AUC is 6.6 times prevalence for ICMEs and 1.5 times for high-speed streams. A random classifier scores 0.5 on ROC-AUC regardless of prevalence, although ROC-AUC depends on the windows labeled positive. Counts of detected events and event POD use the 99th percentile threshold from the validation set; PR-AUC and receiver operating characteristic area under the curve (ROC-AUC) rank scores without a threshold.}
  \label{tab:per_event_type}
  \centering
  \vspace{6pt}
  \resizebox{\textwidth}{!}{%
  \begin{tabular}{lcccccc}
    \toprule
    Event type & Events & Detected & Event POD & PR-AUC$_w$ & Prevalence$_w$ & ROC-AUC$_w$ \\
    \midrule
    Interplanetary shocks (point) & 319 & 79 & 0.248 & 0.071 & 0.032 & 0.655 \\
    High-speed streams (point) & 429 & 46 & 0.107 & 0.064 & 0.044 & 0.598 \\
    ICMEs (interval) & 100 & 69 & 0.690 & 0.350 & 0.053 & 0.838 \\
    \midrule
    All types & 848 & 194 & 0.229 & 0.365 & 0.119 & 0.725 \\
    \bottomrule
  \end{tabular}%
}
\end{table}

\begin{figure}[t]
  \centering
  \includegraphics[width=0.65\textwidth]{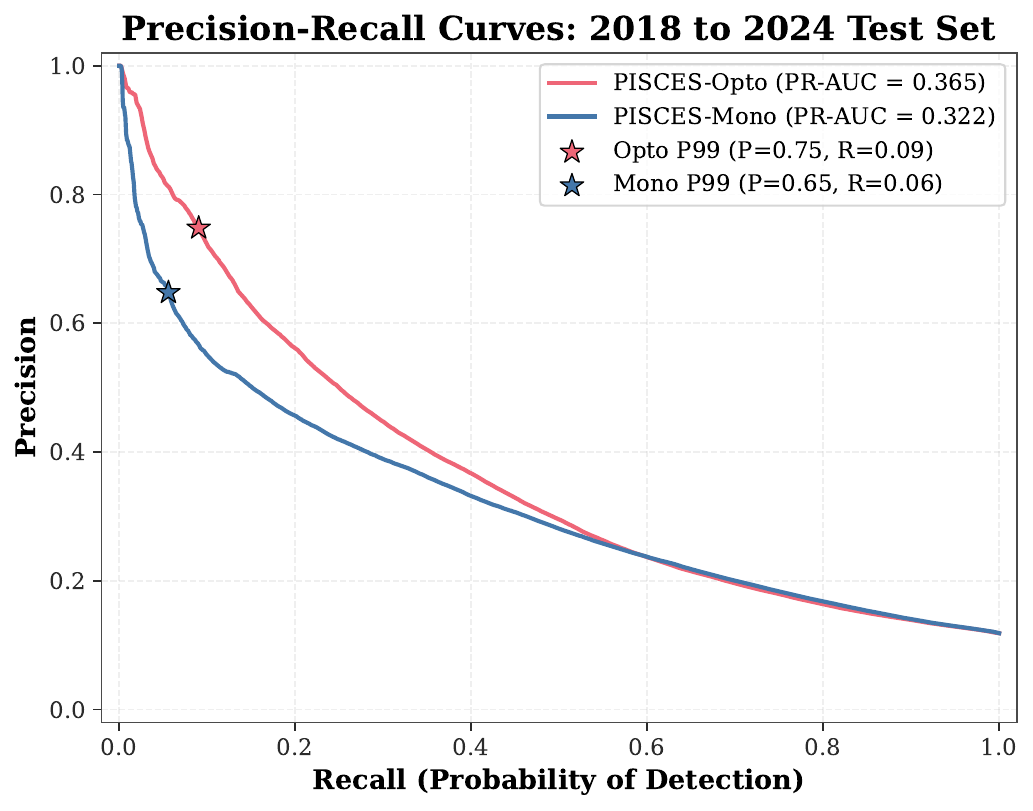}
\caption{Precision-recall curves on the 2018 to 2024 test set for PISCES-Opto and PISCES-Mono (863 catalog events: 328 interplanetary shocks, 433 high-speed streams, 102 interplanetary coronal mass ejections (ICMEs)). Both use the segment procedure and catalog coverage from Table~\ref{tab:main_results}. Stars show test set precision and recall at each model's validation set 99th percentile threshold.}
  \label{fig:pr_curve}
\end{figure}

\begin{figure}[t]
  \centering
  \includegraphics[width=0.95\textwidth]{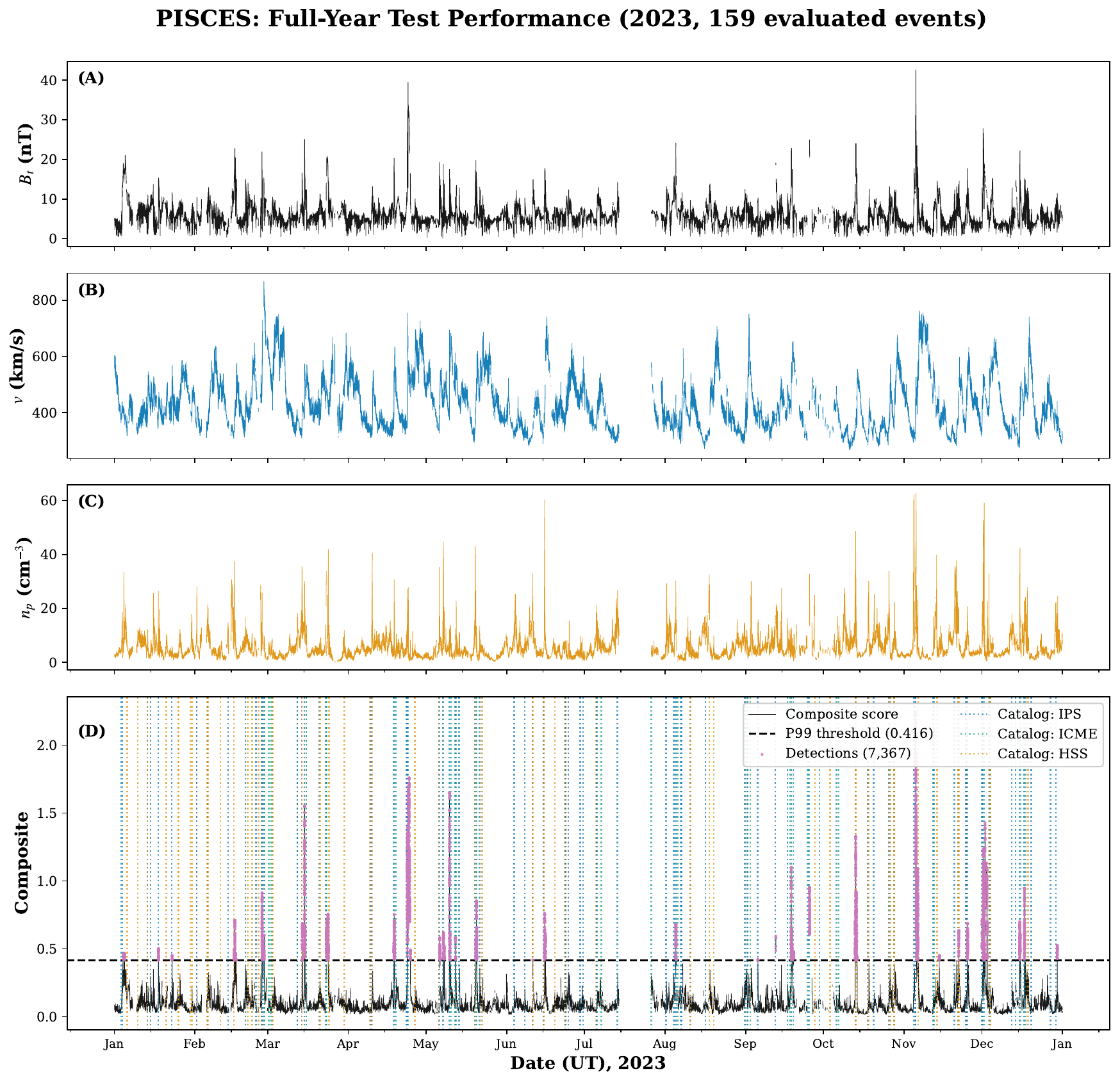}
\caption{PISCES-Mono performance throughout 2023, one year of the 2018 to 2024 test set. The plot shows (A) $B_t$, (B) solar wind speed, (C) proton density, and (D) the PISCES composite anomaly score with its detection threshold at the 99th percentile. Vertical dashed lines mark the starts of DONKI interplanetary shock events (blue), DONKI high-speed stream events (orange), and Richardson \& Cane interplanetary coronal mass ejection (ICME) events (green), using the event colors in Figure~\ref{fig:lead_times}. Many spikes occur near catalog event starts, but the active year has too many markers for a visual correlation check. Some quiet periods remain below threshold. The 2024 test year appears in the Gannon superstorm case study (Figure~\ref{fig:gannon_storm}).}
  \label{fig:test_timeseries}
\end{figure}

Table~\ref{tab:main_results} gives mixed comparisons across methods and metrics. Because each baseline is represented by three runs, these results do not establish that PISCES outperforms every other method. We focus on the score decomposition. Figures~\ref{fig:pr_curve} and \ref{fig:test_timeseries} show precision-recall curves for PISCES-Opto and PISCES-Mono and the complete 2023 time series from the test set, respectively. The figures use the same point estimates as Table~\ref{tab:main_results}.

\clearpage
\subsection{Ablation with Physics Loss Groups}
\label{sec:ablation}

To assess groups of physics losses, we train four model variants for each of five random seeds: $\text{Var}_\mu$~(MSE only, all $\lambda = 0$), $\text{Var}_\gamma$~(MSE with geometry losses $\mathcal{L}_{\text{Bt}}$ and $\mathcal{L}_{\text{parker}}$), $\text{Var}_\theta$~(MSE with thermodynamic losses $\mathcal{L}_{\text{ent}}$, $\mathcal{L}_{\text{pres}}$, $\mathcal{L}_{\text{tv}}$, $\mathcal{L}_{\beta}$, $\mathcal{L}_{\text{dyn}}$), and $\text{Var}_\Sigma$~(complete PISCES, all seven physics losses). All variants use the same hyperparameters, training schedule, and architecture. For each seed, we vary only the configured weights on physics losses. This split separates magnetic field geometry from plasma thermodynamics and tests whether either group changes model behavior.

\begin{table}[H]
\caption{Ablation with physics loss groups on the 2018 to 2024 test set (863 catalog events). Entries are means across five matched training runs. The precision-recall area under the curve (PR-AUC) column also gives its sample standard deviation across the matched training seeds (1, 2, 3, 4, and 42). One script trains all four variants. Within each seed, they differ only in their configured group of physics losses. MSE denotes mean squared error. Average precision is calculated without a threshold. All remaining metrics, including probability of detection (POD) and false alarm ratio (FAR), use the 2016 to 2017 validation set 99th percentile threshold and are then averaged across seeds. Heidke$_w$ is computed from the window-level confusion matrix. The standard deviation measures variability across training runs on this fixed test set, not uncertainty from temporal variation or catalog sampling.}
  \label{tab:ablation}
  \centering
  \vspace{6pt}
  \resizebox{\textwidth}{!}{%
  \begin{tabular}{lccccccc}
    \toprule
    Variant & PR-AUC & POD$_w$ & FAR$_w$ & Precision$_w$ & Heidke$_w$ & Event POD & Episode FAR \\
    \midrule
    $\text{Var}_\mu$: MSE only & $0.291 \pm 0.018$ & 0.057 & 0.393 & 0.607 & 0.086 & 0.193 & 0.486 \\
    $\text{Var}_\gamma$: Geometry only & $0.300 \pm 0.040$ & 0.061 & 0.391 & 0.609 & 0.091 & 0.206 & 0.459 \\
    $\text{Var}_\theta$: Thermodynamic only & $0.309 \pm 0.024$ & 0.063 & 0.360 & 0.640 & 0.096 & 0.208 & 0.441 \\
    $\text{Var}_\Sigma$: Full PISCES & $0.294 \pm 0.033$ & 0.059 & 0.384 & 0.616 & 0.089 & 0.213 & 0.464 \\
    \bottomrule
  \end{tabular}%
}
\end{table}

Table~\ref{tab:ablation} reports means across five matched training runs per variant. The mean average precision is 0.291 for $\text{Var}_\mu$, 0.300 for $\text{Var}_\gamma$, 0.309 for $\text{Var}_\theta$, and 0.294 for $\text{Var}_\Sigma$, while the sample standard deviations are 0.018, 0.040, 0.024, and 0.033. The difference between the best and worst group mean is 0.018, smaller than the range within each single group, which reaches 0.049 for $\text{Var}_\mu$, 0.101 for $\text{Var}_\gamma$, 0.055 for $\text{Var}_\theta$, and 0.085 for $\text{Var}_\Sigma$. The variant with the highest average precision also depends on the seed: $\text{Var}_\gamma$ ranks first once, $\text{Var}_\mu$ once, and $\text{Var}_\theta$ three times. No loss group holds a stable ordering across seeds, so the table is descriptive rather than a test of whether one loss group is better, and we do not attribute the observed differences to particular physics terms.

We also compare training objectives. We completed nine runs: three objectives crossed with three random seeds, 42, 7, and 13. The objectives were PISCES with seven constraints, one based on derived quantities that penalizes inconsistency in reconstructed pressure, entropy, and other derived quantities, and one using reconstruction error alone. Their mean average precisions are 0.304, 0.263, and 0.257, with sample standard deviations 0.055, 0.030, and 0.007. The objective based on derived quantities scores 0.005 higher than the objective using reconstruction error alone on all events pooled and 0.002 to 0.004 higher on each event type separately. The decision rule set before the experiment does not state how large an improvement must be. A later internal note asked for an improvement larger than the spread between runs. These comparisons across three runs do not establish objective superiority. The PISCES-Opto checkpoint, at 0.365, is the best of its three runs here and of all eleven runs later, but its selection criterion was not blind: it was selected based on corrected scores on the test set. The seed used by each run is recorded in the archive path of its checkpoint file, not in the file itself. The checkpoint trained with the PISCES objective for seed 42 was normalized with training set statistics that differ slightly from those of the other eight runs. Averages across three runs therefore combine small archive differences with seed variation.

None of these group or objective means exceeds either of two references with no trained model. An untrained network of the same architecture reaches 0.337, and the window energy score reaches 0.341 by averaging squared standardized inputs without a fitted model. All seven group or objective means fall below both. Four individual trained runs do exceed them: three of the twenty grouped ablation runs and one of the nine objective runs. This is not a consistent pattern. These results do not show that physics terms improve ranking. They support using the fifteen sub-scores to describe physical changes in an event and to provide hints of event types. ICMEs have low proton temperatures, so the expected $T$ and $V$ relationship can fail. Their plasma beta is often much lower than one, and their magnetic field can rotate away from the Parker spiral. These properties correspond to the relevant sub-scores.

\subsection*{Kendall Uncertainty Weighting Failure}
We also tried the homoscedastic task uncertainty weighting described in \cite{kendall2018multi}. Optimization became degenerate because change penalties are zero when their associated derived quantities are constant across a reconstructed window, causing the log variance weights to diverge to $-\infty$ and degrading MSE relative to fixed weights. This degeneracy arises when the weighting objective is applied to a loss that can approach zero. We therefore use fixed weights as described in Section~\ref{sec:physics_loss}.

\paragraph{Hyperparameter Sensitivity:}
Hyperparameter runs from procedures used before hardening were excluded from selection. To select an architecture without 2018 to 2024 data, we performed a rolling-origin study before the 2018 test fold with eight configurations. The configurations varied the physics weight ($pw_{\max}\in\{0,0.10,0.15,0.20,0.30\}$ for PISCES-Opto) and, at $pw_{\max}=0.30$, the architecture (mirror decoder, bottleneck $[32,16,12]$, dilation $[1,2,4]$, or uniform decoder). At this fold resolution, these effects are not separated. Mean PR-AUC across folds varies from 0.429 to 0.445, a spread of 0.016, well below the $\pm 0.07$ standard deviation across folds. The variant trained with MSE alone ($pw_{\max}=0$) ranks highest at 0.445, but the ablation study gives inconsistent results across loss groups (Section~\ref{sec:ablation}). This study does not establish PISCES-Opto as a preferred architecture.

\subsection{Skip Path Attenuation}
\label{sec:skip_interventions}

Anomaly detectors that reconstruct through skip connections can reproduce anomalous structure. This reduces reconstruction error and the resulting anomaly score. We evaluated the PISCES-Opto checkpoint over the 2018 to 2024 test set with skip contribution scales of 1.0, 0.5, 0.25, and 0.0. At scale 0.0, skip feature channels are zeroed before concatenation with decoder features, while the $1\times1$ fusion convolutions remain active. Setting skip contribution to zero raises test average precision from 0.365 to 0.406 and increases the composite score scale by approximately three times. Reconstructions worsen throughout the test set, while the distinction between anomalous and quiet windows increases. This intervention removes direct skip feature contributions at inference. Full skips can partially reconstruct anomalies, at a cost of about 0.04 average precision.

\begin{table}[H]
\caption{Skip path attenuation in the test period from 2018 to 2024. For each trained checkpoint, we chose the attenuation scale with the highest average precision in the validation period from 2016 to 2017. Entries across eleven runs give means and standard deviations. The table includes three models retrained without skip connections. They also omit fusion parameters. Committed artifacts include the aggregate and evaluation records for each run; raw scale scans for only three of the eleven trained checkpoints are retained. This retrospective study was designed after inspecting results on the test set. Selection uses validation average precision only.}
  \label{tab:skip_interventions}
  \centering
  \vspace{6pt}
  \resizebox{\textwidth}{!}{%
  \begin{tabular}{llc}
    \toprule
    Configuration & Attenuation selection & Test average precision \\
    \midrule
    PISCES-Opto checkpoint, full skips & none & 0.365 \\
    PISCES-Opto checkpoint, selected scale & scale 0 & 0.406 \\
    \midrule
    Mean across eleven runs ($\pm$ sd), full skips & & $0.306 \pm 0.029$ \\
    Mean across eleven runs ($\pm$ sd), selected scale & per checkpoint & $\text{\textbf{0.376}} \pm \text{\textbf{0.020}}$ \\
    \midrule
    Models retrained without skip connections (3 runs) & not applicable & $0.228 \pm 0.001$ \\
    \bottomrule
  \end{tabular}%
  }
\end{table}

We selected the attenuation scale separately for each run from its 2016 to 2017 validation data without tuning it against the test score. For the released checkpoint, validation average precision rises monotonically with attenuation, from 0.224 at full skips to 0.328 at scale 0. The validation criterion selects scale 0 for this checkpoint. Across all runs, selection chooses scale 0 for three runs and scale 0.25 for eight. All eleven runs improve when attenuation is selected on validation, and their mean changes from $0.306 \pm 0.029$ at full skips to $0.376 \pm 0.020$ with selected attenuation. Removing the skips triples the composite score scale, so the operating points need new validation thresholds at the 99th percentile, from 0.239 with full skips to 0.871 with skip attenuation.

Attenuation selected on validation data improves window ranking while reducing event coverage. Test average precision is 0.406, with a 95\% CI of 0.345 to 0.462 from monthly blocks. Window POD$_w$ rises from 0.090 to 0.110 and window FAR$_w$ falls from 0.252 to 0.210, while event POD falls from 0.229 to 0.154. The attenuated detector concentrates flagged windows in long events such as ICME interiors.

No single attenuation scale works equally well for every run. The intervention was designed after inspecting results on the test set, so it remains retrospective rather than a blind test of skip attenuation. The two operating points use separate 99th percentile thresholds from validation and do not match the numbers of alarm windows or alarm episodes.

The untrained runs test whether scale selection is a byproduct of random initialization. Across the four scales, untrained validation average precision changes by at most $6.4\times10^{-4}$. Validation therefore provides little evidence for a preferred untrained scale. The untrained comparison alone does not explain the much larger trained gain.

Retraining without skip connections changes the training architecture, whereas attenuation changes inference for an already trained checkpoint. Three retrained models reach average precision between 0.227 and 0.230, lower than PISCES-Opto with full skips and with inference attenuation. The runs retrained without skip connections also omit the parameters that fuse skip features, which brings the parameter count from 12,555 down to 11,571, so this comparison does not isolate why skipless training performs worse. Within the tested setups, training with strong skip dropout and then omitting skip contributions at inference scores higher than training without skip connections.

\subsection{An Untrained Network}
\label{sec:untrained_control}

The preceding results combine contributions from trained weights and from the rest of the system. Architecture, training set normalization, composite formula, and thresholding procedure remain fixed whether weights are trained or reinitialized. To compare trained and reinitialized weights, we reinitialize the checkpoint's learned parameters and batch normalization running statistics, then evaluate four independent random initializations. During inference, those layers then pass inputs through unchanged, whereas the trained checkpoint has running variances far from one. Two additional runs estimate the effect of setting those statistics from training-set windows with random weights.

Table~\ref{tab:untrained_control} compares untrained and trained results. With full skips, untrained runs reach average precision $0.337 \pm 0.005$ across four draws, compared with $0.306 \pm 0.029$ across eleven trained runs. Two runs with batch-normalization statistics estimated from training windows reach $0.339 \pm 0.013$. Ten of the eleven trained runs score below the untrained mean. We compare the absolute difference in group means across all 1365 assignments of the four untrained and eleven trained runs. The untrained group has the higher mean at $p = 0.054$, marginal rather than decisive.

The eleven runs do not all share one lineage. Restricting to the homogeneous subset omits the selected checkpoint and two runs from another experiment, leaving eight trained runs from lineage \texttt{seeds\_2026-07-31}. The eight remaining trained runs are $0.031$ below the untrained mean, with $p = 0.008$. This result depends on the lineage restriction, so it does not resolve the full sample comparison. Training's contribution to ranking remains unproven while skips are at full strength.

The untrained decoder output is near zero. In two of the four standard untrained runs, decoder output energy is $0.327\%$ and $0.409\%$ of input energy on 7,581 finite 2018 windows advanced 60 minutes at a time, while the trained checkpoint's output is $96.4\%$ of input energy. When decoder output is near zero in standardized coordinates, the MSE sub-scores approximate input energy in their respective channel groups. For those runs, average precision for reconstruction only and for the full composite differs by at most 0.0002, against 0.019 for the trained checkpoint in Section~\ref{sec:weight_analysis}. Four sub-scores use inputs directly, and two compare inputs with reconstructions. The final row of Table~\ref{tab:main_results} computes window energy directly and reaches average precision of 0.341, above ten of the eleven trained runs.

Two comparisons favor trained weights, with important qualifications. First, skip attenuation selected on validation data increases average precision for every trained run by $+0.070 \pm 0.028$ (sign test $p = 0.001$). Untrained runs gain only $+0.0001 \pm 0.0003$. Their validation scores change by at most $6.4 \times 10^{-4}$ across the four scales, while trained runs span 0.023 to 0.157 across their four scales. Validation selects full skips twice and complete removal twice for the untrained runs, so those choices provide little evidence of a preferred scale. Second, trained runs have event POD $0.201 \pm 0.016$, compared with $0.103 \pm 0.001$ for untrained runs at their separately derived operating points. Every trained run is higher than every untrained run, and the separation is significant at the floor of an exact permutation test over 1365 label permutations, $p = 7\times10^{-4}$. Alarm counts differ as well: trained runs have 2.6 times as many alarm episodes, and event POD normalized by windows and episodes changes direction. This does not isolate a training effect. Average precision evaluates window rankings, while event detection uses a threshold, and the two can differ at a positive rate of 11.9\%. The energy row in Table~\ref{tab:main_results} illustrates this distinction: it reaches 0.341 average precision but has event POD 0.101.

With attenuation selected on validation data, the same eleven runs reach $0.376 \pm 0.020$ in average precision. Every run exceeds the untrained mean, and the difference remains significant at an exact permutation $p = 0.004$ against the four untrained runs with default batch normalization statistics. The mean also exceeds the tuned tree baseline from Section~\ref{sec:discussion}, 0.373. Table~\ref{tab:untrained_control} reports this comparison across eleven runs alongside the untrained runs.

The decomposition ordering also differs between trained and untrained runs. We rank the fifteen component scores by effect size and compare these rankings across runs. The three runs trained without physics terms have mean pairwise Spearman correlation $0.937$. Their mean correlation with the eleven runs trained with physics terms is $0.913$, and with the six untrained runs is $0.242$. Models trained with different objectives give similar component rankings. The ordering is consistent with trained weights. Physics terms in the objective alone do not explain it.

\begin{table}[H]
\caption{Untrained networks. These runs use the same architecture, input normalization, composite, thresholding, and evaluation process as the trained rows, but their weights were never trained. Default-statistic runs leave batch normalization at its initial running statistics, while runs with batch-normalization statistics estimated from training windows use randomly initialized weights. Attenuation is selected on the validation set for both trained and untrained checkpoints. Each run and inference configuration derives its own 99th percentile threshold from its 2016 to 2017 distribution of validation scores. Average precision is calculated without a threshold, whereas event POD uses that separately derived operating point.}
  \label{tab:untrained_control}
  \centering
  \vspace{6pt}
  \begin{tabular}{lccc}
    \toprule
    Configuration & Runs & Average precision & Event POD \\
    \midrule
    Trained, full skips & 11 & $0.306 \pm 0.029$ & $0.201 \pm 0.016$ \\
    Trained, attenuation selected on validation & 11 & $\text{\textbf{0.376}} \pm \text{\textbf{0.020}}$ & $0.157 \pm 0.007$ \\
    \midrule
    Untrained, full skips & 4 & $0.337 \pm 0.005$ & $0.103 \pm 0.001$ \\
    Untrained, batch normalization estimated & 2 & $0.339 \pm 0.013$ & $0.114 \pm 0.003$ \\
    Untrained, attenuation selected on validation & 4 & $0.337 \pm 0.005$ & $0.103 \pm 0.001$ \\
    \bottomrule
  \end{tabular}
\end{table}

\subsection{Trailing Median Smoothing and Estimated Warning Leads}
\label{sec:warning}

The offline procedure described above uses a centered 15-minute median, including roughly seven minutes of future scores. For the warning analysis, we use a trailing 15-minute median for both configurations, so each output uses the current score and preceding scores only. We derive each 99th percentile threshold from validation scores after applying that filter. For the checkpoint initialized with seed 42, full skips score average precision of 0.362 (95\% CI 0.303 to 0.416 from a bootstrap over blocks defined by calendar month), compared with 0.365 under centered smoothing. Attenuation selected on validation data scores 0.403 (CI 0.343 to 0.459), compared with 0.406 under centered smoothing. Window and event operating characteristics differ by no more than 0.005. We use OMNI data shifted to the bow shock as described above.

We evaluate warning timing in the OMNI stream, which is sampled once per minute and has timestamps representing predicted arrival at the bow shock nose. An L1 observation at wall clock time $\tau$ appears at ${\tau + \Delta}$ in OMNI, where $\Delta$ is the delay from L1 to the bow shock. Subtracting $\Delta$ estimates the spacecraft observation time associated with an OMNI timestamp. It does not establish when the complete input window was available or when an operational alert could have been issued. Shock timing requires a further correction because shocks propagate faster than trailing plasma. For 80 shocks with abrupt L1 onsets in this test period, measured delay from L1 to the ground has a median of 46 minutes. A trained timing model gives 47 minutes when supplied with mean input values \cite{baumann2021timing}. Flow speed gives 62 minutes. We fit measured transit times with $delay = a + b/V$, using $a = 2.9$ minutes. The intercept includes magnetosheath and ground travel times, and the residual median absolute deviation is about 6 minutes. In the zero-intercept inverse-speed sensitivity model restricted to 71 storm sudden commencements, the fitted coefficient changes by about 1\%. Shock margins reference predicted ground onset. Sudden-commencement leads reference observed ground onset.

OMNI publishes a time shift from L1 to the bow shock, and the choice of shift affects lead estimates. The sudden-commencement analysis uses the median published OMNI time shift from five minutes before through 15 minutes after each event where available. Of 130 events, 125 provide an OMNI value; the remaining five have neither an OMNI value nor an estimate of flow transit time over a fixed distance and are dropped. At sudden-commencement times, the estimate using a fixed distance has a median about 4 minutes longer than OMNI's shift, with about 13 minutes of scatter between events. Because each minute of the difference becomes a minute of estimated warning lead, using OMNI's column reduces leads by about four minutes. The eligible count rises from 109 to 112 because changing the shift changes the pairing interval and its coverage requirement, which is at least half of that interval.

First, interplanetary shocks. DONKI times are recorded at L1 and OMNI times at the bow shock, so we recenter each shock on its front, identified in OMNI data by the largest joint speed and magnetic field jumps near the catalog time. This identifies fronts for 196 of 328 shocks. The others fall in data gaps or have no identifiable front. An alarm timestamp qualifies if it falls between 30 minutes before the identified front and 3 hours afterward. Only alarm timestamps within this interval are counted. For the checkpoint initialized with seed 42, full skips at the validation 99th percentile threshold match 71 out of the 196 anchored shocks, with median margin $+16.6$ minutes (interquartile range (IQR) = $-11$ to $+32$, 69\% positive). With attenuation selected on validation data, the model matches 41 out of the 196 anchored shocks, with median margin $+12.6$ minutes (59\% positive). Table~\ref{tab:warning} gives margin ranges across three runs. It omits coverage of anchored shocks because the normalization check below reverses its direction.

Second, we compare alarms with independently observed sudden-commencement (SC) events from the International Service on Rapid Magnetic Variations \cite{curto2007evolution}. The event set includes storm sudden commencements (SSCs) and positive sudden impulses. Negative impulses are excluded. It uses ground magnetometer observations independent of the L1 measurements, the DONKI and Richardson \& Cane catalogs, and model outputs. Observatori de l'Ebre maintains the list, which is definitive through 2022 and provisional for 2023 to 2024. Ground magnetometers record the response to a shock or pressure front at the magnetosphere. They supply an onset time independent of the L1 measurements. We pair each sudden commencement with the closest eligible alarm episode, using each event and episode at most once. For this analysis, an alarm episode joins alarm timestamps separated by at most 15 minutes; this differs from the contiguous alarm episode count used for episode FAR. An episode start is eligible from two hours before the event through that event's transit delay afterward. The pairing interval permits only nonnegative estimated leads, and the warning criterion requires a strictly positive lead. Coverage and estimated lead magnitude are the reported outcomes. Expanding the two-hour lookback to three or six hours changes the number of matches by no more than three. For the released checkpoint with seed 42, 39 (35\%; 95\% CI 26 to 44\% from event bootstrap) of 112 events eligible for sudden-commencement coverage are paired with full-skip alarms whose estimated leads before ground onset are positive, with median lead $+26$ minutes (IQR 16 to 37). Attenuation selected on validation data covers 20 (18\%; CI 11 to 25\%) of 112 eligible events, with median lead $+19$ minutes. Table~\ref{tab:warning} reports ranges across all three runs, including 32 to 39 of 112 at full skips. The timing study uses 380 paired L1 shock and sudden-commencement observations \cite{baumann2021timing}. Across three matched runs, after smoothing with a trailing median, pairing alarms with independently observed storm sudden commencements and positive sudden impulses gives median estimated leads before ground impact of 24 to 28 minutes for 29 to 35\% of events with full skips, against 18 to 19 minutes for 18 to 20\% under attenuation selected on validation. The attenuated configuration has higher average precision but pairs alarms with fewer sudden commencements.

We compare both warning endpoints with the untrained models of Section~\ref{sec:untrained_control}. After smoothing with a trailing median, each of the four untrained runs matches 12 of the same 112 sudden commencements, compared with 32 to 39 for the three trained runs. For anchored shocks, the counts are 36 to 38 for untrained runs and 63 to 71 for trained runs. The trained runs produce 2.6 times as many alarm episodes, 469 on average versus 178, which accounts for most of the 3.0 coverage ratio for sudden commencements and all of the 1.8 ratio for shocks. Normalizing by alarm episode changes the interpretation: trained runs have a 14\% higher sudden-commencement match rate than untrained runs (0.074 to 0.081 versus 0.067 to 0.069), while coverage of anchored shocks reverses. Every trained run matches fewer shocks per alarm episode than every untrained run (0.138 to 0.160 versus 0.201 to 0.211), with a ratio of trained to untrained matches per alarm episode of 0.70 and a 44\% margin in favor of the untrained runs. Both comparisons hold at $p = 0.029$ on an exact permutation test over the 35 possible permutations of three trained and four untrained runs, the smallest attainable value from this design. Coverage of anchored shocks does not establish a training advantage because its direction reverses between normalization by alarm windows and alarm episodes. We therefore withdraw the coverage endpoint for anchored shocks.

For sudden commencements, untrained median leads are 11.1 to 12.2 minutes, compared with 23.9 to 27.9 minutes for trained runs. For shocks, untrained median margins are $-6.6$ to $-3.0$ minutes, so the median matched alarm occurs after predicted ground onset. Trained median margins are $+16.6$ to $+18.1$ minutes. The two tests use different subsets of matched events, so they do not establish a paired timing effect. At full skips, trained runs have higher sudden-commencement matches per alarm episode but fewer matches of anchored shocks per alarm episode.

\begin{table}[H]
\caption{Warning summary across three runs with a trailing 15-minute median score filter, 2018 to 2024. SC denotes the 112 events eligible for coverage, comprising storm sudden commencements and positive sudden impulses. Average precision after the trailing 15-minute median is the mean $\pm$ sample standard deviation across matched runs with seeds 7, 13, and 42. The sample standard deviation describes variation across these heterogeneous runs, not seed variation alone. The remaining cells are descriptive ranges of values across runs. Each configuration uses its own 99th percentile threshold from validation. In the column for selected attenuation, seed 42 uses scale 0 and seeds 7 and 13 use scale 0.25. Shock margins reference ground onset predicted by the empirical delay model, whose residual median absolute deviation is about six minutes. Operational latency is not subtracted. SC leads reference onset times from the ground magnetometer catalog. The ranges are not confidence intervals or paired treatment effects.}
  \label{tab:warning}
  \centering
  \vspace{6pt}
  \begin{tabular}{lcc}
    \toprule
    & Full skips & Selected attenuation \\
    \midrule
    Test average precision after trailing median & $0.301 \pm 0.055$ & $0.378 \pm 0.025$ \\
    SC events with leading alarm, run range & 32 to 39 of 112 & 20 to 22 of 112 \\
    Median estimated SC lead, range (min) & $+24$ to $+28$ & $+18$ to $+19$ \\
    Median shock margin, range (min) & $+17$ to $+18$ & $+5$ to $+13$ \\
    \bottomrule
  \end{tabular}
\end{table}

\begin{figure}[htbp]
  \centering
  \includegraphics[width=\textwidth]{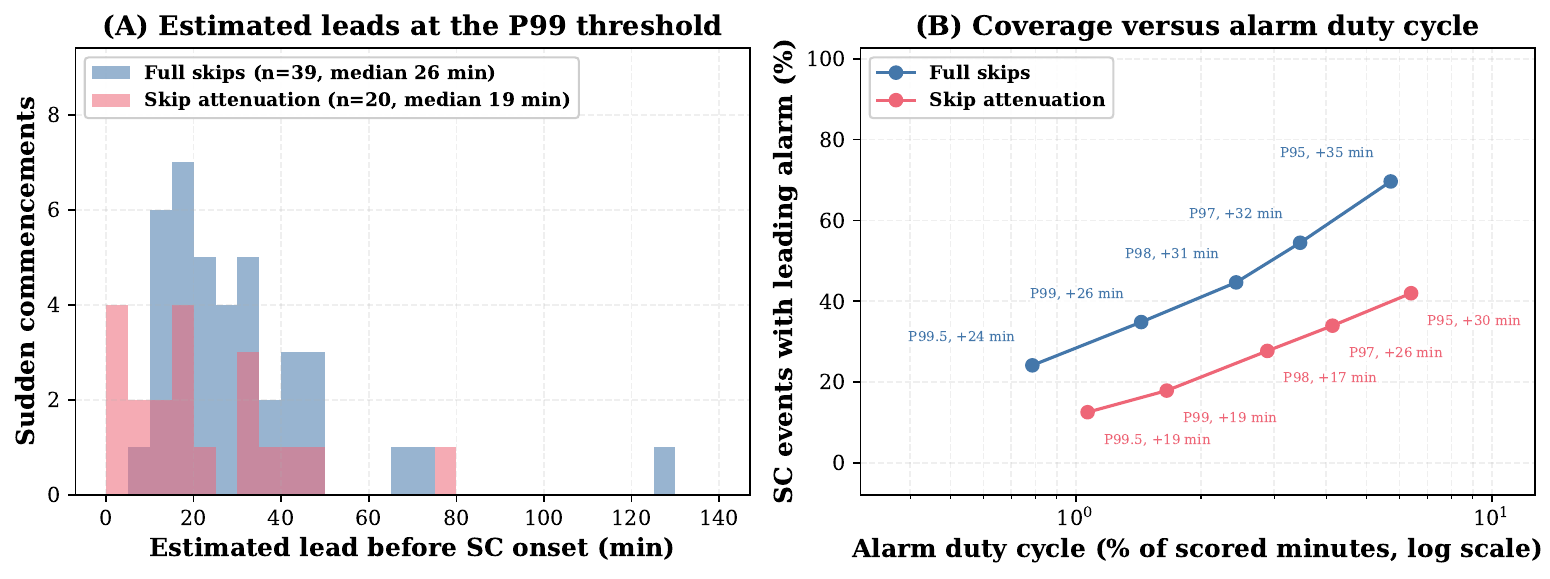}
\caption{Warning behavior with a trailing 15-minute median. \textbf{(A)}~Estimated lead before onset of the paired ground event at the 99th percentile validation threshold. Each lead adds the median of OMNI's published time shift over the interval from five minutes before through 15 minutes after the event. It does not establish sample availability or operational latency. A trained timing model gives a 47-minute delay for mean input values. This provides context, not a bound on individual alarm leads \cite{baumann2021timing}. An alarm episode beginning before the L1 shock can yield a longer estimated lead, and one pair reaches $+125$ minutes. \textbf{(B)}~Coverage of storm sudden commencements and positive sudden impulses versus alarm duty cycle across thresholds set at different validation percentiles, with the median lead estimate at each point. These are retrospective estimates from the development data. P95, P97, P98, P99, and P99.5 denote the 95th, 97th, 98th, 99th, and 99.5th validation percentiles. Section~\ref{sec:warning} gives the caveats.}
  \label{fig:warning_leads}
\end{figure}

Full skips cover more sudden commencements at every tested validation percentile. Figure~\ref{fig:warning_leads} sweeps the validation threshold from its 95th percentile to its 99.5th percentile, changing both sudden-commencement coverage and alarm duty cycle. At P95, the setting with full skips associates 78 of 112 sudden-commencement events (70\%) with a median alarm lead of $+35$ minutes at 5.7\% duty cycle, while at the 99.5th percentile it still associates 27 of 112 events at 0.8\% duty cycle.

The raw ACE analysis tests transfer from OMNI to archived ACE Level 2 measurements in spacecraft time. It uses ACE magnetometer data sampled every 16 seconds and Solar Wind Electron Proton Alpha Monitor (SWEPAM) data sampled every 64 seconds, bins both to one-minute cadence, and uses OMNI normalization statistics. We apply no additional adjustment across spacecraft or time shift to the bow shock. Model weights, score weights, and OMNI normalization statistics remain fixed. Standard preprocessing fills short gaps and excludes windows containing longer gaps. The 99th percentile threshold is calculated from raw ACE validation scores from 2016 to 2017 after applying the same trailing 15-minute median. The ACE proton temperature is the radial component. In the comparison for the quiet day of 15 April 2020, the median ACE minus OMNI temperature difference is about 14\% of the median OMNI temperature. Availability of Level 2 SWEPAM data through 2024-07-09 limits the analysis. On the covered subset of approximately 1.12 million minutes with 452 eligible events and a positive rate of 0.115, the raw ACE analysis achieves average precision of 0.415 (95\% confidence interval (CI) 0.341 to 0.484 from a bootstrap over blocks defined by calendar month) at full skips and 0.475 (CI 0.393 to 0.547) with skip attenuation.

On the exact overlap, both score series are evaluated against the same labels at 948,623 common timestamps, with equal prevalence and no additional timestamp shift. Raw ACE average precision is 0.415 and 0.462, compared with 0.369 and 0.434 for OMNI scores after smoothing with a trailing median, for differences of $+0.045$ (95\% confidence interval 0.028 to 0.061) and $+0.028$ (CI 0.011 to 0.045). The quoted confidence intervals describe this comparison without an additional timestamp shift. To assess timebase sensitivity, moving OMNI score timestamps 40 minutes earlier raises OMNI average precision by about 0.02. Subtracting shifted OMNI average precision from raw ACE average precision gives numerical differences of $+0.026$ and $+0.011$, but the time shift drops some timestamp pairs. These comparisons use different sets of timestamps and do not isolate timestamp alignment from coverage differences. The ACE results use fixed model and score weights with the same preprocessing and smoothing with a trailing median.

Within the ACE analysis period, 83 of the 112 sudden-commencement events have fewer than 60 scored minutes in the two hours before onset. The remaining 29 meet this coverage criterion. Two have a preceding full-skip alarm, and none has a preceding alarm with attenuation. The OMNI warning results use a different set of events eligible for coverage.

\begin{figure}[htbp]
  \centering
  \includegraphics[width=\textwidth]{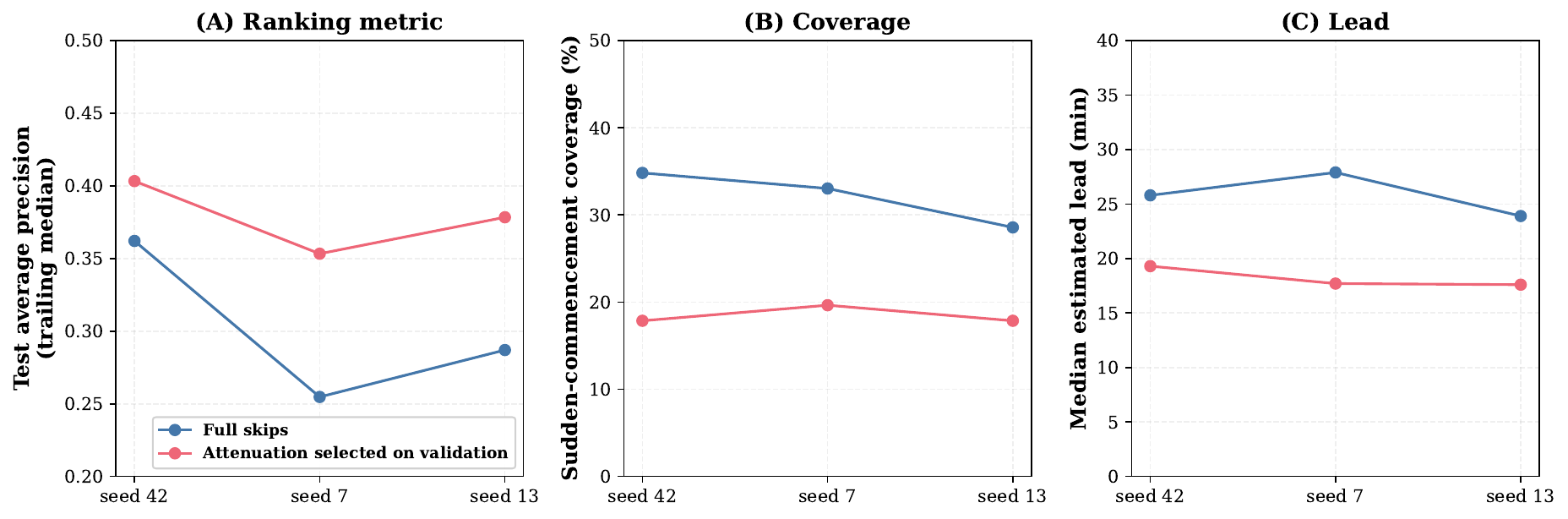}
\caption{Variation in warning endpoints across runs. \textbf{(A)}~Average precision after the trailing 15-minute median by training run and configuration. Scores range from 0.25 to 0.40. \textbf{(B)}~Sudden-commencement coverage and \textbf{(C)}~median estimated lead time for the same experiments. Full skips cover 29 to 35 percent of events, with medians of 24 to 28 minutes. Attenuation covers fewer events, with medians of 18 to 19 minutes. The procedure and data are the same as in Table~\ref{tab:warning}.}
  \label{fig:warning_seeds}
\end{figure}

We repeated the warning analysis on two additional runs and selected attenuation on validation for each. Average precision after the trailing 15-minute median ranges from 0.255 to 0.362 with full skips and from 0.353 to 0.403 with attenuation. Across all three runs, full-skip sudden-commencement coverage is 32 to 39 of 112, with median lead times of $+24$ to $+28$ minutes and shock margins of $+17$ to $+18$ minutes. Using the trailing median lowers average precision by about 0.003. The raw ACE analysis keeps each run's model and score weights fixed, derives only its ACE validation 99th percentile threshold, and has average precision from 0.31 to 0.47. Figure~\ref{fig:warning_seeds} shows these results. Full skips cover more sudden commencements than attenuation in every run, but operating thresholds and alarm counts differ. Median estimated sudden-commencement leads are shorter with attenuation. Each median represents its own subset of matched events and does not define a pairwise attenuation effect. Shock margins with attenuation are $+5$ to $+13$ minutes. The warning ranges show variation across runs as well as differences between configurations.

Attenuation gives higher average precision, which has no threshold. Full skips associate more sudden commencements with alarms, although their alarm counts and durations differ. This is the distinction described in Section~\ref{sec:skip_interventions} between ranking performance and event POD.

\subsection{Cross Validation Across Solar Cycle Phases}
\label{sec:temporal_robustness}

A single split between training and test data can overestimate performance, so we evaluated three alternatives: a rolling-origin split before the test years, a rotating monthly split, and a stress test blocked by solar rotation. We first checked the split definitions because one variant is unusable. The retrained monthly variant trains on neighbors of the test month, while the variant that keeps the trained weights fixed scores two test months twice and omits two others.

The new design splits 2005 to 2024 into sequential blocks of one, two, or four solar rotations. A solar rotation lasts about 27 days, the recurrence period of corotating wind \cite{schrijver2010evolving}. One unused rotation borders each test block, which is longer than any catalog event; events touching a boundary are discarded. Every fold retrains PISCES-Opto on the remaining data and evaluates it as in Table~\ref{tab:main_results}. Of 460 folds, 80 are degenerate and omitted: 62 folds with test blocks spanning one rotation, 13 with blocks spanning two, and 5 with blocks spanning four contain no retained catalog event.

The remaining 380 folds average 0.265 (203 folds, s.d.\ 0.19), 0.273 (118 folds, s.d.\ 0.16), and 0.264 (59 folds, s.d.\ 0.14) PR-AUC for blocks spanning one, two, or four rotations. Block size has little effect in this aggregate, whereas solar cycle phase does. Blocks spanning one rotation near the cycle 24 maximum, from 2012 to 2015, average 0.35 across 47 folds. Blocks near the 2019 to 2020 minimum average 0.12 across 24 folds. Scores also increase with the number of events in a block because average precision depends on the base rate. Each fold uses a different training subset and evaluates a short test block. We therefore compare it with the mean across the eleven runs, $0.306 \pm 0.029$. The single checkpoint's 0.365 is not a comparable reference. The difference between 0.265 and 0.306 may reflect differences in training set composition, evaluation on short blocks, or training run variation. This analysis does not separate these effects.

An earlier sweep over 460 folds used physics weight 0.2 despite being labeled pw30, followed the earlier scoring setup, and reported a mean of 0.376 across valid folds. These corrected values supersede that report. Dividing a time series into contiguous blocks and validating across them is a common response to serial correlation \cite{bergmeir2012use,lopezdeprado2018advances}. Other ICME studies have used both yearly cross validation \cite{rudisser2022automatic,rudisser2026arcane} and chronological holdout \cite{nguyen2019automatic}. The designs answer different questions: blocks test the training setup across solar cycle phases, while holdout tests performance on an unseen future.

\FloatBarrier
\subsection{Score Decomposition Analysis}
\label{sec:decomposition}

\begin{figure}[htbp]
  \centering
  \includegraphics[width=0.9\textwidth]{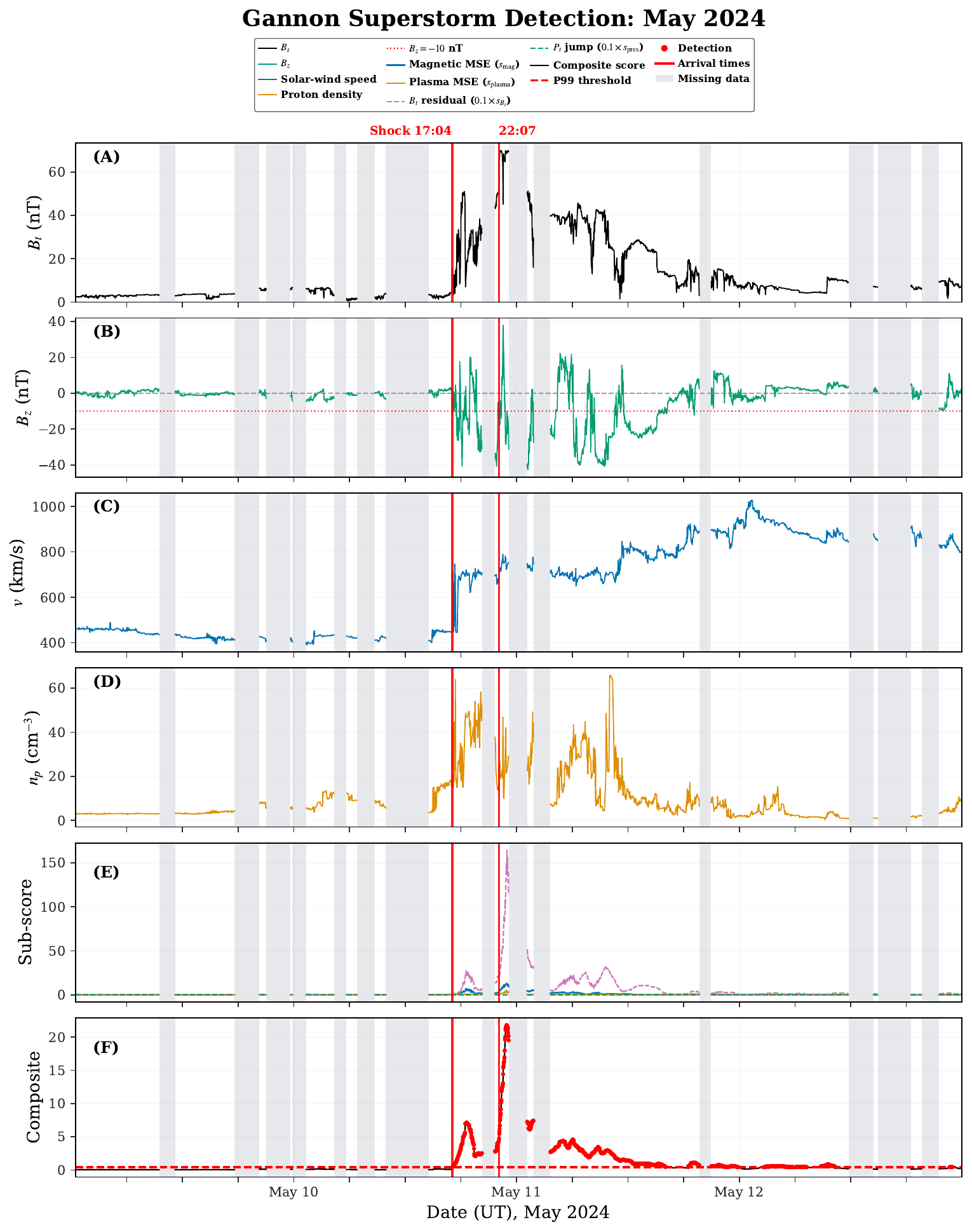}
\caption{PISCES-Mono anomaly detection during the May 2024 Gannon superstorm, the test period's largest geomagnetic storm, rated G5 by the National Oceanic and Atmospheric Administration (NOAA). The top four subplots show denormalized, preprocessed OMNI parameters in physical units. The bottom subplots show the magnetic and plasma mean squared error (MSE) sub-scores, the $B_t$ consistency residual, the total pressure jump, and the composite score, with the detection threshold at the 99th percentile shown by a red dashed line. In subplot~(E), the traces for the $B_t$ residual and $P_t$ jump are scaled to one tenth for legibility. Times are Universal Time (UT). Red vertical lines mark the interplanetary coronal mass ejection (ICME) forward shock at 17:04~UT in the OMNI frame and a time marker at 22:07~UT. DSCOVR observed the forward shock at L1 at 16:34~UT and a possible later ICME arrival at 21:36~UT \cite{tulasiram2024gannon}. These precede the OMNI markers by 30 and 31 minutes, respectively. OMNI shifts upstream measurements to predicted arrival at the bow shock nose \cite{papitashvili2020omni}. PISCES shows elevated anomaly scores throughout the storm, peaking near the shock interval.}
  \label{fig:gannon_storm}
\end{figure}

\begin{figure}[p]
  \centering
  \includegraphics[width=0.88\textwidth]{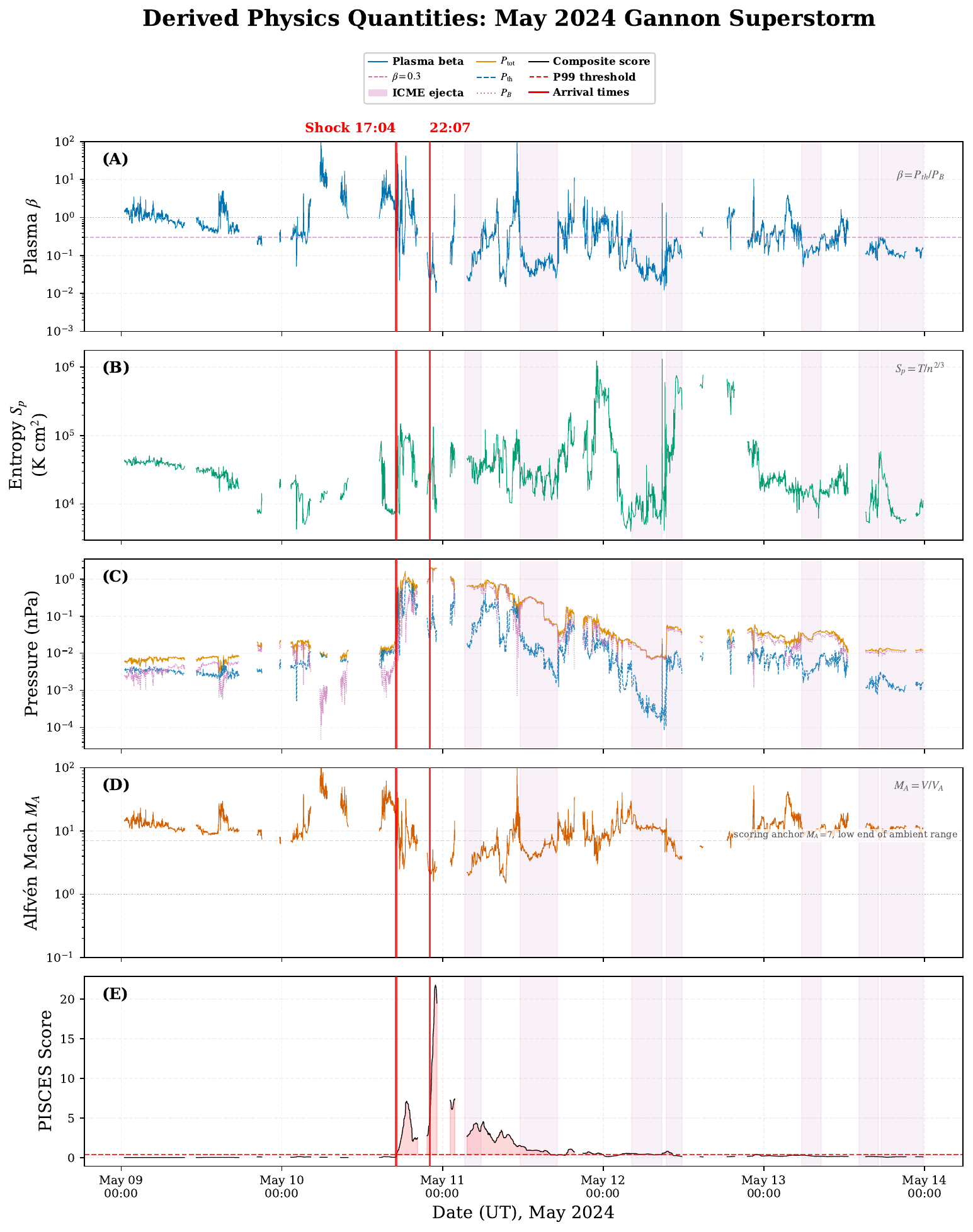}
\caption{Derived physical quantities from PISCES-Mono sub-score calculations during the May 2024 Gannon superstorm (May~9 to 14). \textbf{(A)}~Plasma $\beta = P_{\text{th}}/P_B$ is below 0.3 during the interplanetary coronal mass ejection (ICME) encounter. Purple shading marks intervals where $\beta$ remains below 0.3 for more than two hours, a magnetic ejecta (ME) indicator derived from magnetic field and plasma measurements. It does not use the catalog interval. Figures~\ref{fig:insitu_gannon} and \ref{fig:hodogram_gannon} use the same criterion and data, so all three figures cover the same intervals. The plots remain on the scoring grid because they show the model inputs. \textbf{(B)}~Proton entropy proxy $S_p = T_p / n_p^{2/3}$. \textbf{(C)}~Total pressure estimate, the sum of proton thermal and magnetic pressure, peaks near 2~nPa, about 256 times the median before the storm. It remains tens of times larger during the first day after the shock (median 79 times, interquartile range 34 to 99) before settling to a few times baseline on the second day. \textbf{(D)}~Alfv\'{e}n Mach number $M_A = V/V_A$ approaches unity during the storm, with a minimum near 1.2. This minimum occurs between the shaded periods. Within the low-$\beta$ ejecta, $M_A$ remains closer to 2. \textbf{(E)}~PISCES composite score with the 99th percentile threshold. Periods without valid model windows remain blank in the plot. Times are Universal Time (UT). Vertical red lines mark the interplanetary coronal mass ejection forward shock at 17:04~UT in the OMNI frame and a time marker at 22:07~UT. DSCOVR observed the forward shock at L1 at 16:34~UT and a possible later ICME arrival at 21:36~UT \cite{tulasiram2024gannon}. These precede the plotted markers by 30 and 31 minutes, respectively.}
  \label{fig:physics_case_study}
\end{figure}

\begin{figure}[htbp]
  \centering
  \includegraphics[width=\textwidth]{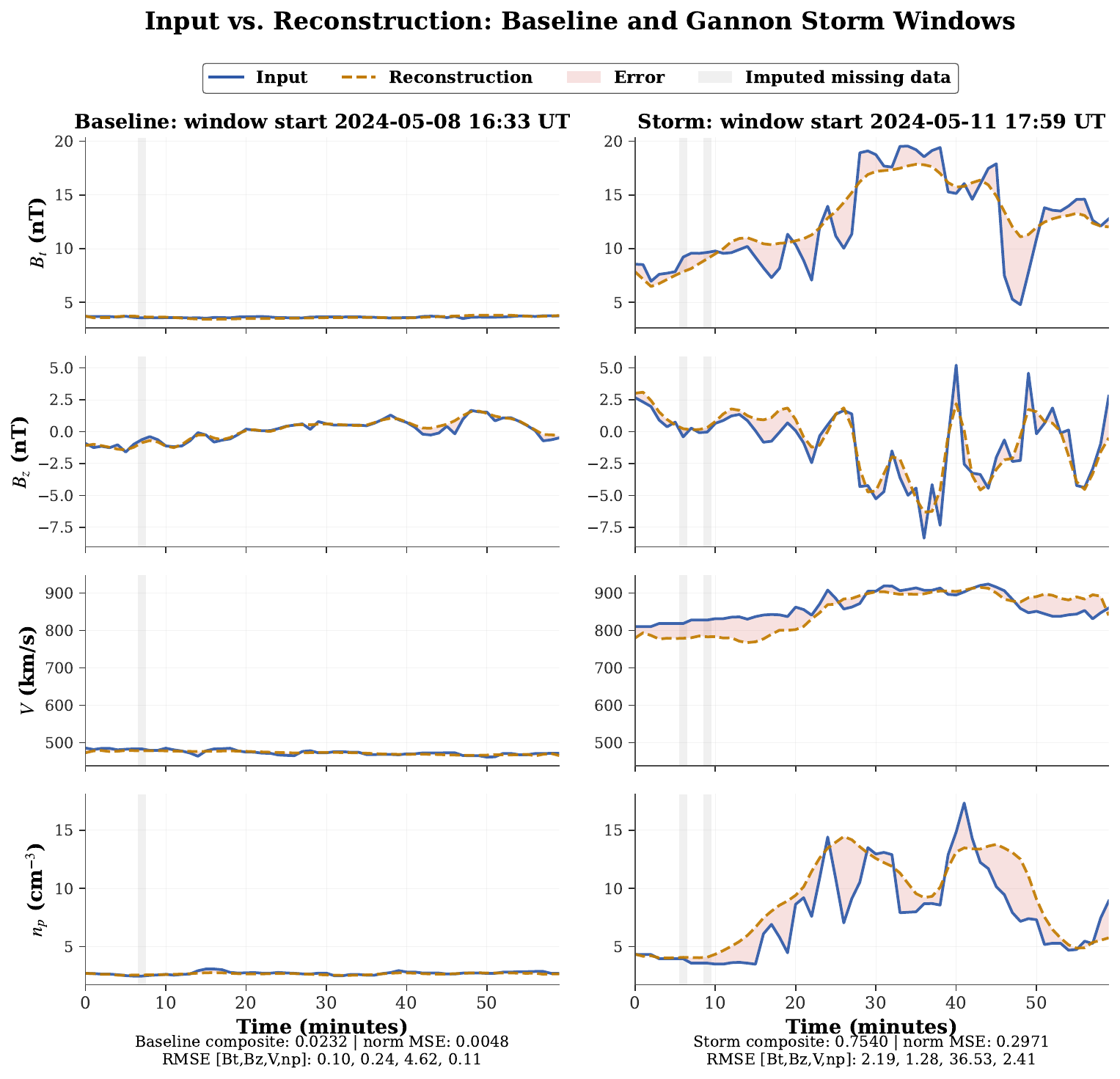}
\caption{Input and PISCES-Opto reconstruction for two 60-minute solar wind windows. All times are Universal Time (UT). The reference window with a low anomaly score on the left starts at 2024-05-08 16:33~UT, and the Gannon superstorm window with a high anomaly score on the right starts at 2024-05-11 17:59~UT. Four example channels, $B_t$, $B_z$, $V$, and $n_p$, are shown in physical units. Gray shading marks OMNI samples that were missing and imputed during preprocessing. The reconstruction follows the window with a low anomaly score but deviates in the storm window, especially in magnetic and plasma channels. The composite score rises from $0.023$ to $0.754$, and normalized mean squared error (MSE) rises from $0.0048$ to $0.297$. The subplot annotations report root mean square errors (RMSEs) by channel in the displayed physical units.}
  \label{fig:reconstruction}
\end{figure}

The decomposition shows which quantities contribute to each window's anomaly score. Figure~\ref{fig:gannon_storm} shows the sub-scores during the May 2024 Gannon superstorm, a severe geomagnetic storm during solar cycle 25 with terrestrial effects measured by ground magnetometers \cite{demichelis2025gannon} and satellite drag \cite{parker2024gannon}. Figure~\ref{fig:reconstruction} compares inputs and reconstructions for windows with different anomaly levels. Figure~\ref{fig:physics_case_study} presents physical quantities used in the training losses and anomaly scores during this event.

\begin{figure}[htbp]
  \centering
  \includegraphics[width=\textwidth]{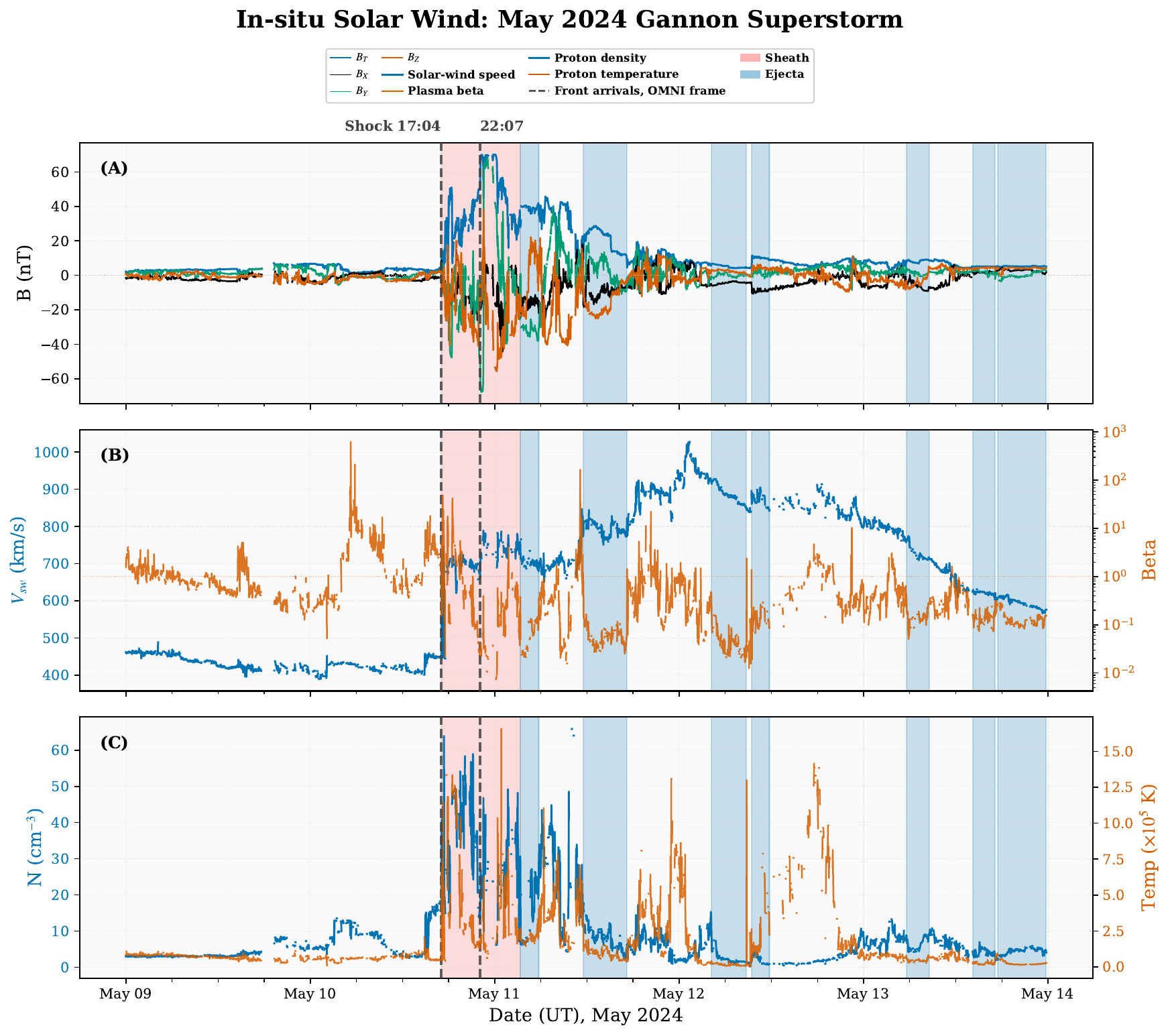}
\caption{In-situ solar wind measurements during the May~2024 Gannon superstorm (May~9 to 14) from NASA OMNI data sampled once per minute. \textbf{(A)}~Interplanetary magnetic field (IMF) magnitude $B_T$ (blue) and components $B_X$ (black), $B_Y$ (green), and $B_Z$ (orange), in nT. \textbf{(B)}~Solar wind speed $V_{sw}$ (km/s, left axis) and $\beta = P_{\text{th}}/P_B$ (right axis, log scale). \textbf{(C)}~Proton density $N_p$ (cm$^{-3}$, left axis) and proton temperature $T_p$ ($10^5$~K, right axis). Times are Universal Time (UT). Vertical dashed lines mark the interplanetary coronal mass ejection forward shock at 17:04~UT in the OMNI frame and a time marker at 22:07~UT. DSCOVR observed the forward shock at L1 at 16:34~UT and a possible later ICME arrival at 21:36~UT \cite{tulasiram2024gannon}. These precede the plotted markers by 30 and 31 minutes, respectively.}
  \label{fig:insitu_gannon}
\end{figure}

\begin{figure}[p]
  \centering
  \includegraphics[width=\textwidth, height=0.70\textheight, keepaspectratio]{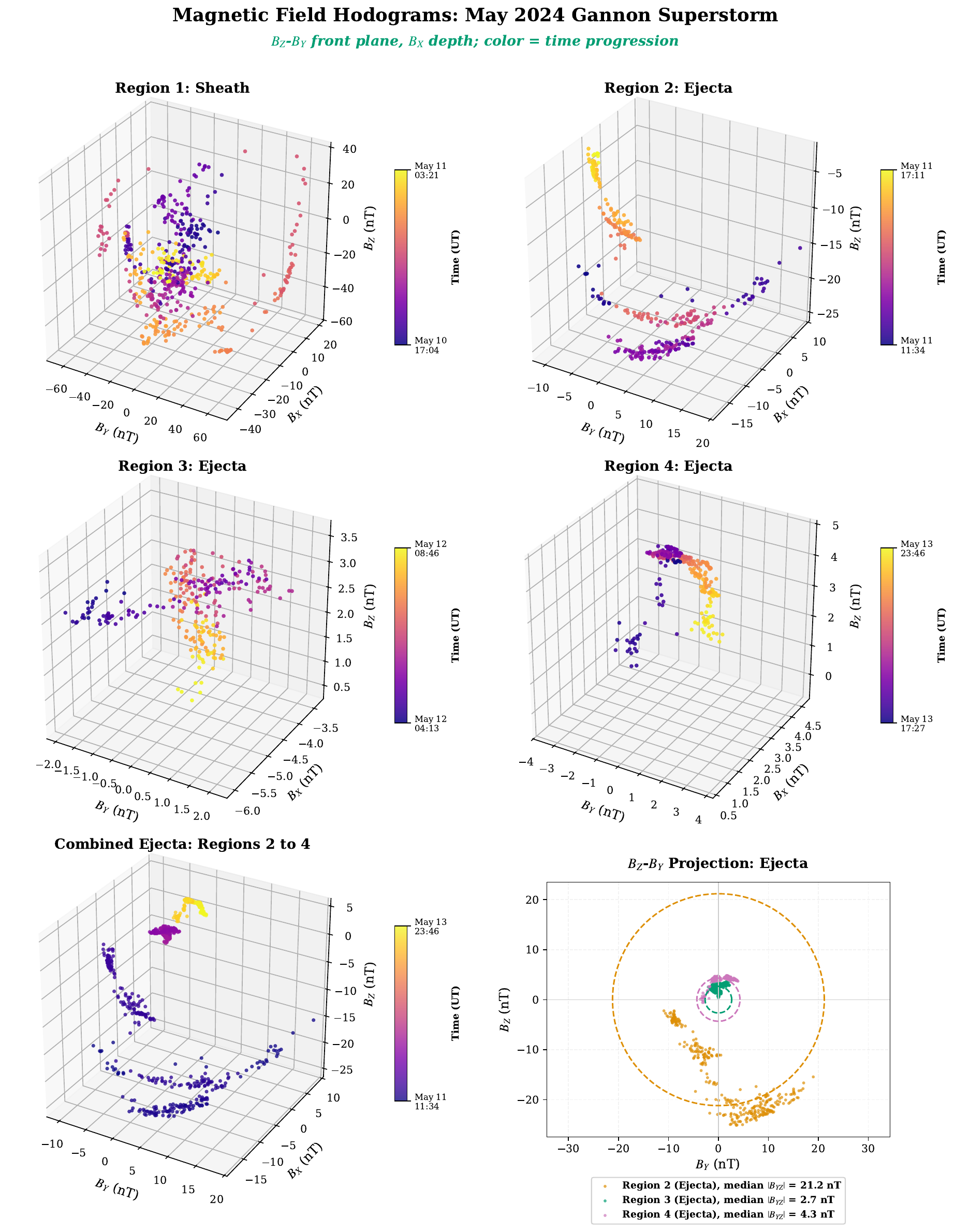}
\caption{Three-dimensional magnetic field hodograms, colored by time, for four solar wind regions during the May 2024 Gannon superstorm, from NASA OMNI data sampled once per minute. Each plot shows $B_Y$ (front), $B_X$ (depth), and $B_Z$ (vertical) in nT, with elapsed time encoded from dark purple to light yellow. Times are in Universal Time (UT). Regions use the same criteria and data as Figure~\ref{fig:insitu_gannon}: $\beta<0.3$ for at least 2~h, allowing up to 30~min of data loss with at least 50\% coverage in each interval. The sheath extends from the shock to the first qualifying low-$\beta$ interval, which spans 03:21 to 05:44~UT on May~11. This shorter interval is not among the three longest intervals plotted as Regions~2 to 4. \textbf{Region~1: Sheath}: from shock arrival in the OMNI frame at 17:04~UT on May~10 to 03:21~UT on May~11, about 10.3~h, with disordered field and a root mean square (RMS) field increment over 1 minute of about 11~nT, 4.6 times that of Region~2. Tulasi Ram et al.\ \cite{tulasiram2024gannon} attribute the composite ICME to CMEs that interacted en route to Earth. The sheath label is operational and does not identify the interval as a conventional sheath ahead of a single ICME. \textbf{Region~2: Ejecta}, May~11 11:34 to 17:11~UT (about 5.6~h): a coherent arc in the $B_Z$ versus $B_Y$ plane at a median transverse field of about 21~nT, the most distinct rotation of the three. \textbf{Region~3: Ejecta}, May~12 04:13 to 08:46~UT (about 4.5~h): an incomplete, diffuse rotation in a weaker field (about 3~nT). \textbf{Region~4: Ejecta}, May~13 17:27 to 23:46~UT (about 6.3~h): a similarly weak field (about 4~nT) later in the eruption sequence, with little net rotation. \textbf{Bottom left}: the three plotted ejecta intervals together. \textbf{Bottom right}: the $B_Z$ versus $B_Y$ plot with dashed circles marking each region's median $|B_{YZ}|$.}
  \label{fig:hodogram_gannon}
\end{figure}

Figure~\ref{fig:insitu_gannon} shows the in-situ solar wind data for this event. The sequence of shock, sheath, ejecta, and recovery from Figure~\ref{fig:mhd_phase} is visible. Dashed lines mark the shock transition, where velocity and density jump abruptly, followed by a turbulent, compressed sheath with high $|B|$. Figure~\ref{fig:hodogram_gannon} shows three-dimensional magnetic field hodograms for all four regions. The Parker spiral score measures departures from the expected field angle, and the rotation score measures rapid changes in field direction.

\begin{figure}[htbp]
  \centering
  \includegraphics[width=0.85\textwidth]{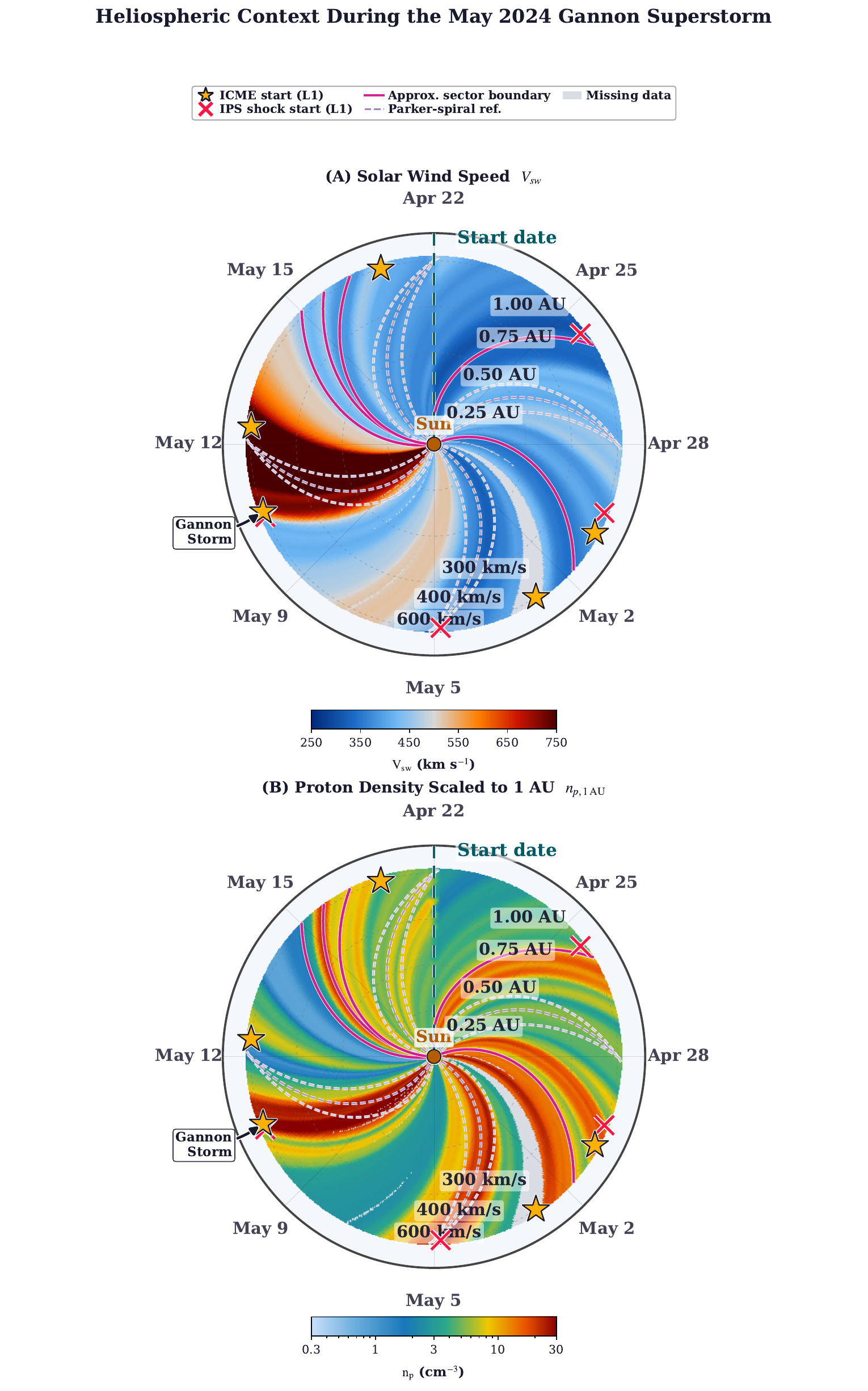}
\caption{Heliospheric context during the May 2024 Gannon superstorm. Valid near-Earth OMNI samples are mapped toward the Sun at their recorded solar wind speed \cite{nolte1973large}. Polar angle shows synodic Earth/L1 Carrington date from 22 April to 19 May 2024, and radius shows heliocentric distance in astronomical units (AU). Date labels are near-Earth observation times: a sample observed at time $t$ maps to radius $r$ at $t-(1-r)\,\mathrm{AU}/V_{sw}$. The dashed radial line gives the 22 April reference date. \textbf{(A)}~Solar wind speed. \textbf{(B)}~Proton density scaled to 1~AU. Magenta curves mark estimated sector boundaries, where smoothed OMNI $B_x$ changes sign under the same mapping. Dashed Parker spirals at 300, 400, and 600 km/s use Parker's spiral geometry \cite{parker1958dynamics} at the Carrington rotation rate and constant speed. They are geometric references, not magnetic-connectivity or plasma-flow paths. Light gray triangles mark missing OMNI data. Stars show all five Richardson \& Cane interplanetary coronal mass ejection (ICME) onsets in the plot window, and crosses show four of twelve DONKI interplanetary shocks (IPS).}
  \label{fig:heliospheric_map}
\end{figure}

\begin{figure}[htbp]
  \centering
  \includegraphics[width=\textwidth]{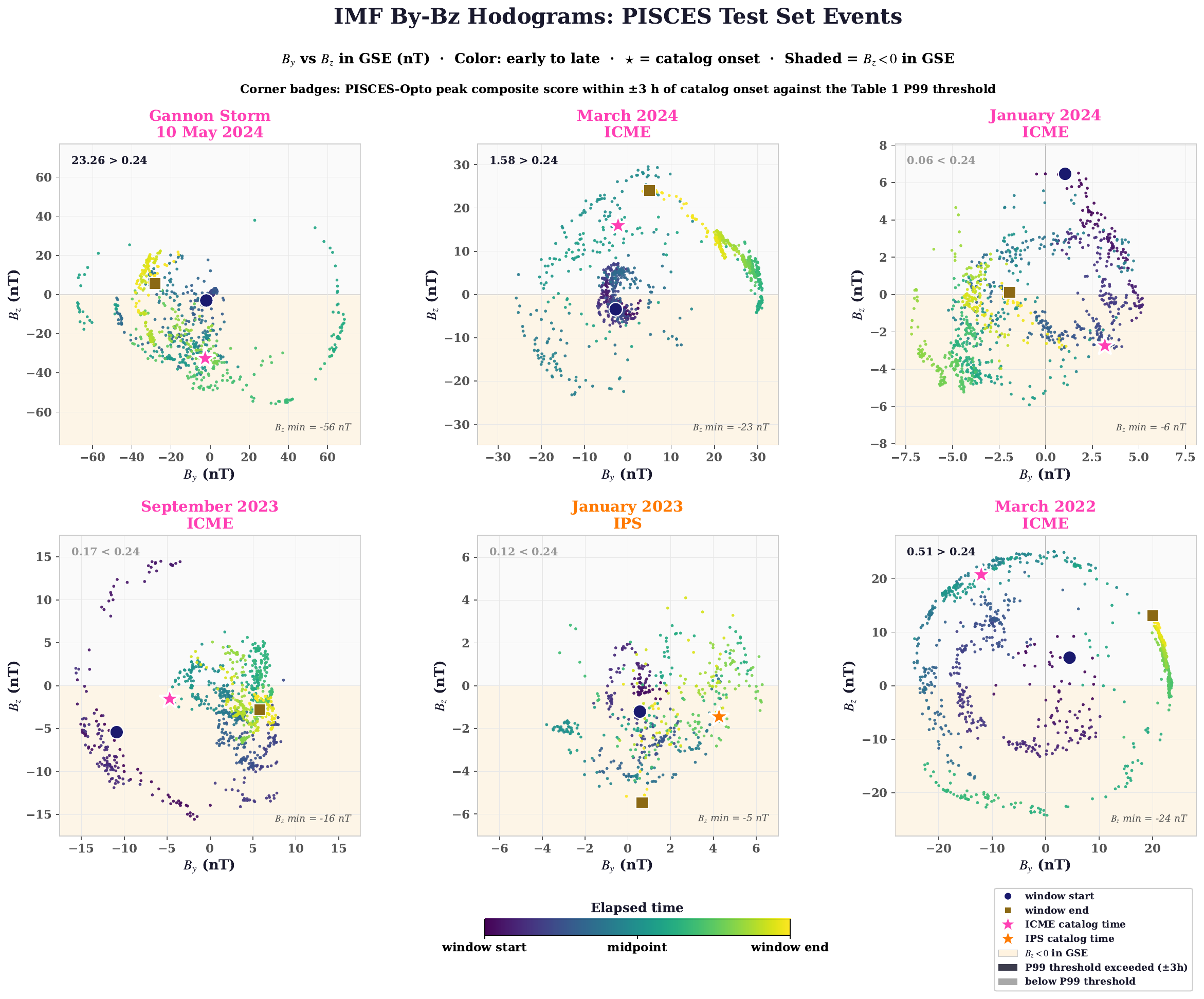}
\caption{Interplanetary magnetic field (IMF) $B_y$ versus $B_z$ hodograms in geocentric solar ecliptic (GSE) coordinates for six selected 2018 to 2024 test set events in OMNI data sampled once per minute, with elapsed time encoded from dark purple to yellow. The circle marks the window start, the square marks the window end, and the star, colored by event type, marks the sample closest to catalog onset. The warm amber region marks $B_z < 0$ in GSE. Geomagnetic coupling also depends on field orientation relative to Earth's magnetic field. The badge reports the PISCES-Opto checkpoint's maximum composite anomaly score within $\pm3$~h of catalog onset, relative to the 99th percentile validation threshold. The plot centers its window on onset. For interplanetary coronal mass ejection (ICME) labels in Table~\ref{tab:main_results}, the evaluated interval extends from 3 hours before the catalog start to 3 hours after the catalog end, so corner values are not directly comparable with detection statistics. Most ICMEs (magenta) produce a coherent rotation arc in the $B_y$ versus $B_z$ plane. Strong field rotations characterize flux rope structure \cite{bothmer1998structure, lepping1990magnetic}. These hodograms show GSE components without rotation into a flux rope frame. The January~2024 rotation has a different form. The January~2023 interplanetary shock (IPS) event (orange) has a clear shock compression field jump. A clear signature does not guarantee a high corner score because the validation threshold comes from the full window score distribution.}
  \label{fig:clock_angle}
\end{figure}

Figure~\ref{fig:heliospheric_map} places the Gannon superstorm in a heliospheric context by mapping near-Earth OMNI data into heliospheric coordinates. The map is qualitative and does not solve the MHD equations or establish exact magnetic connectivity. It maps valid near-Earth samples to earlier dates assuming constant speed and marks approximate sector boundaries. Figure~\ref{fig:clock_angle} shows $B_y$ versus $B_z$ hodograms for six selected test set events. ICME flux ropes trace coherent rotation arcs, while interplanetary shocks appear as abrupt turns between components. Corner values give the checkpoint's peak composite score within $\pm3$~h of catalog onset, relative to the 99th percentile threshold.

\begin{table}[t]
\caption{Score statistics for PISCES-Mono on the 2018 to 2024 test set. Sub-scores are computed for each window. MSE denotes mean squared error. Means and maxima use the full test set. ``Source'' identifies the quantity used to compute each sub-score: R = reconstruction, I = input, and R$-$I = reconstruction residual relative to input. The $B_t$ consistency residual has the largest ratio of maximum to mean. Its numerical scale also depends on its units and definition.}
  \label{tab:score_stats}
  \centering\footnotesize
  \vspace{6pt}
  \begin{tabular}{llcc}
    \toprule
    Sub-score & Source & Mean & Max \\
    \midrule
    \multicolumn{4}{l}{\itshape Reconstruction error} \\
    $s_{\text{mag}}$ (magnetic MSE) & R & 0.036 & 13.3 \\
    $s_{\text{plasma}}$ (plasma MSE) & R & 0.020 & 5.41 \\
    \midrule
    \multicolumn{4}{l}{\itshape Physics constraint scores} \\
    $s_{\text{Bt}}$ ($B_t$ consistency residual) & R & 3.584 & 1644.2 \\
    $s_{\text{ent}}$ (entropy jump) & R & 0.121 & 1.63 \\
    $s_{\text{pres}}$ (total pressure jump) & R & 0.099 & 1.22 \\
    $s_{\text{tv}}$ ($T$ versus $V$ relation deviation) & R & 0.594 & 4.90 \\
    $s_{\text{parker}}$ (Parker spiral) & R & 0.351 & 1.03 \\
    $s_{\beta}$ (beta jump) & R & 0.222 & 1.81 \\
    $s_{\text{dyn}}$ (dynamic pressure jump) & I & 0.276 & 3.06 \\
    \midrule
    \multicolumn{4}{l}{\itshape Scoring with no corresponding training loss} \\
    $s_{M_A}$ (Alfv\'{e}n Mach deviation) & R & 0.380 & 2.44 \\
    \midrule
    \multicolumn{4}{l}{\itshape Input-only scores} \\
    $s_{\text{rot}}$ (magnetic field rotation) & I & 0.141 & 1.30 \\
    $s_{\text{mf}}$ (mass flux jump) & I & 0.264 & 3.05 \\
    $s_{\text{tv,in}}$ (input $T$ versus $V$ relation deviation) & I & 0.427 & 3.56 \\
    \midrule
    \multicolumn{4}{l}{\itshape Physics residuals} \\
    $s_{\text{res,ent}}$ (entropy residual) & R$-$I & 0.044 & 0.61 \\
    $s_{\text{res,pres}}$ (pressure residual) & R$-$I & 0.043 & 0.88 \\
    \midrule
    Composite & & 0.083 & 21.8 \\
    \bottomrule
  \end{tabular}
\end{table}

Table~\ref{tab:score_stats} shows the mean and maximum of each sub-score over all 2,969,144 PISCES-Mono test windows. The composite is a sum with constant weights: reconstruction sub-scores have unit weights, while physical and derived sub-scores have weights from 0.002 to 0.005.

\subsection{Physics Loss Interpretation}
\label{sec:weight_analysis}

The ablation in Section~\ref{sec:ablation} does not support attributing aggregate accuracy to individual physics terms. At this sample size, training variability is larger than the observed differences between loss groups. The seven terms still describe physical conditions in each window: magnetic consistency, pressure and entropy structure, deviation from the temperature and velocity relationship, plasma beta, dynamic pressure, and Parker spiral geometry. Each sub-score measures one such condition, while reconstruction error retains physical meaning.

\begin{figure}[htbp]
  \centering
  \includegraphics[width=0.85\textwidth]{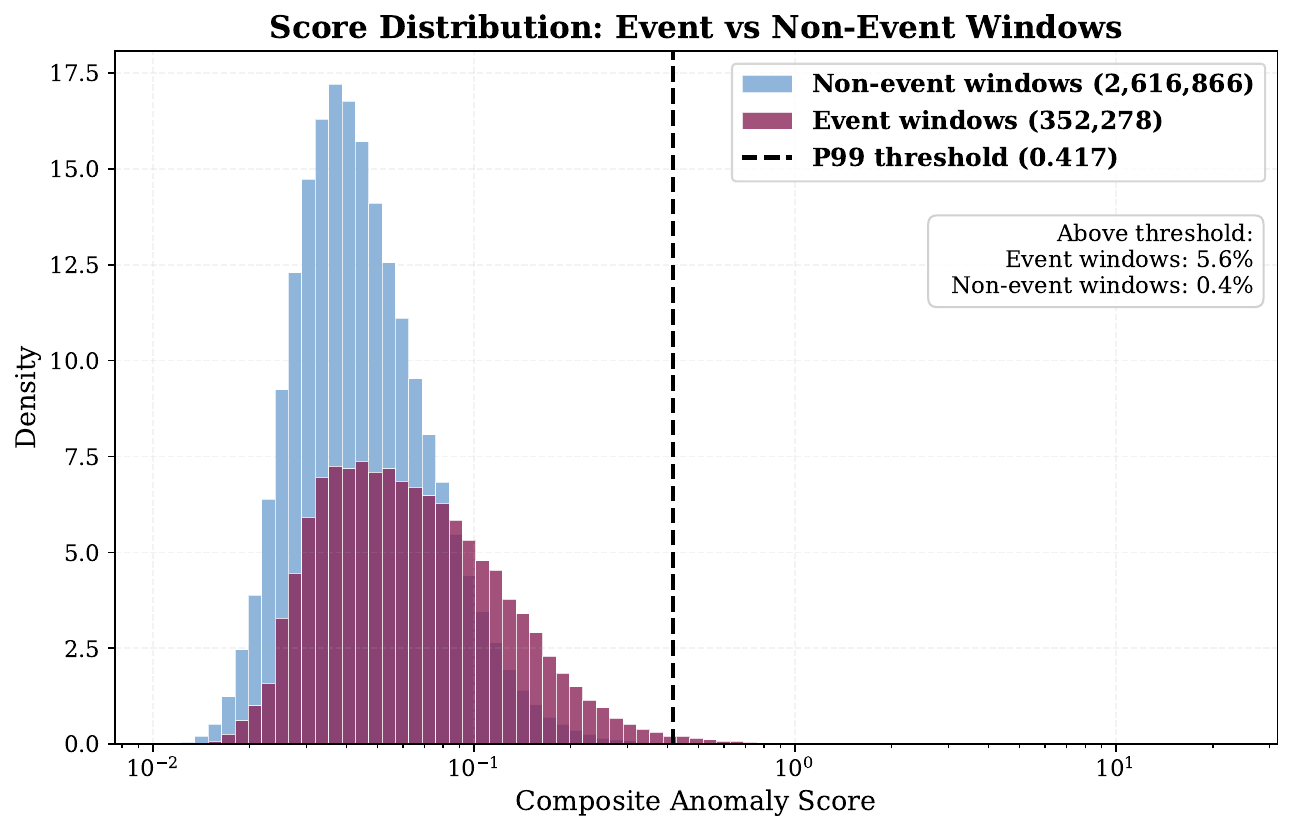}
\caption{Distribution of PISCES-Mono composite anomaly scores for windows from events and windows without events on the 2018 to 2024 test set. Event labels follow Table~\ref{tab:main_results}: shocks and high-speed streams use $\pm3$~h around onset, while interplanetary coronal mass ejections (ICMEs) use the catalog interval extended by 3 hours at each end. Scores are smoothed within continuous coverage, and the catalog is restricted to scored events. On the logarithmic axis, event window scores lie to the right of scores from windows without events. Scores exceed the 99th percentile validation threshold in 5.6\% of event windows and 0.4\% of windows without events. The 5.6\% is the window probability of detection (POD). The 0.4\% is the window false positive rate, which has a different denominator from false alarm ratio (FAR) in Table~\ref{tab:main_results}.}
  \label{fig:score_distribution}
\end{figure}

\begin{figure}[htbp]
  \centering
  \includegraphics[width=0.7\textwidth]{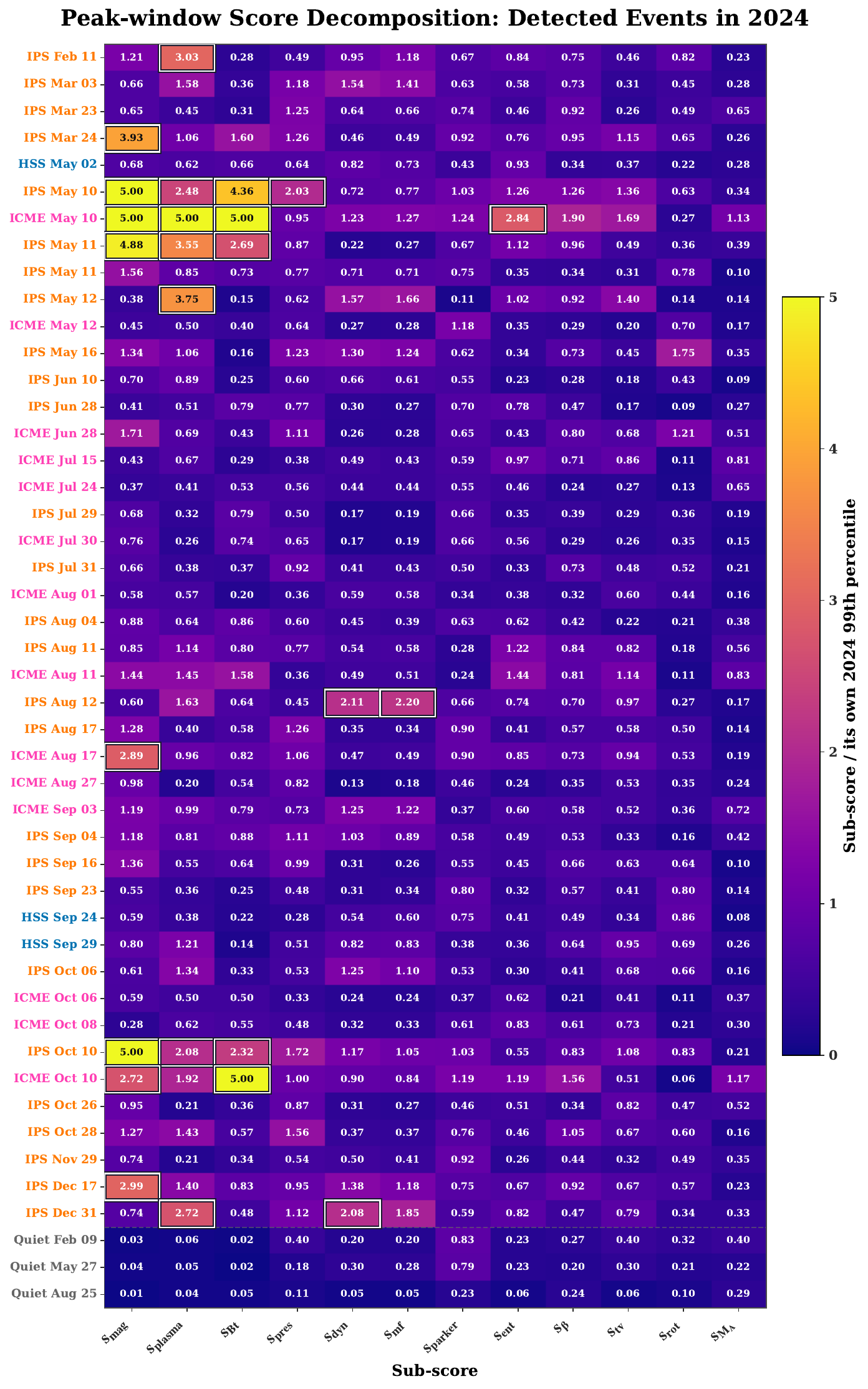}
\caption{PISCES-Mono score decomposition for 2024, the final test year and the year of the Gannon superstorm. Each event row shows the window with the peak composite score for a detected catalog event; events whose peak score does not exceed the threshold have no row. Matching follows Table~\ref{tab:main_results}: $\pm3$~h around onset for interplanetary shocks and high-speed streams, and the interval from 3 hours before start to 3 hours after end for ICMEs. Rows are sorted by time and labeled by type and date: interplanetary shocks (orange), high-speed streams (blue), interplanetary coronal mass ejections (ICMEs, magenta), and quiet intervals (gray). This figure shows individual peak windows, not event-type averages. Three quiet intervals, labeled ``Quiet'' in the bottom rows, are sampled at temporal quartiles from windows without events whose composite scores lie below the 10th percentile over all windows. The columns show 12 of 15 sub-scores, omitting $s_{\text{tv,in}}$ and residual scores $s_{\text{res,ent}}$ and $s_{\text{res,pres}}$. Each score is normalized by its 99th percentile among all scored 2024 windows, capped at 5, and ordered by decreasing mean response. Cells show normalized values. Values above 2.0 have white borders outlined in navy. Quiet intervals have reconstruction and jump responses close to zero, while $s_{\text{pres}}$, $s_{\text{tv}}$, and $M_A$ deviation retain nonzero quiet baselines. Event rows show responses consistent with their physical signatures. Because event rows use windows with each event's highest composite score and quiet rows use windows with low scores, this contrast is partly constructed. Reconstruction scores $s_{\text{mag}}$ and $s_{\text{plasma}}$ dominate all event types.}
  \label{fig:score_decomposition}
\end{figure}

Figure~\ref{fig:score_distribution} shows the difference between scores in event windows and windows without events. The long right tail of scores in event windows extends past the 99th percentile, while scores in windows without events peak at lower values. Figure~\ref{fig:score_decomposition} displays the sub-score responses for detected events in 2024.

The reconstruction terms $s_{\text{mag}}$ and $s_{\text{plasma}}$ dominate the composite, and the composite defines detection. They are therefore large in detected event rows and small in quiet rows by construction. Physics sub-scores carry little composite weight. Relative to their respective 99th percentiles, mean $s_{\text{Bt}}$ and $s_{\text{pres}}$ values over the three quiet intervals are approximately 0.03 and 0.23. The corresponding means for the three event types range from 0.34 to 1.24 and 0.48 to 0.94, respectively.

This is not independent evidence. Each sub-score contributes to the composite, and selecting the window with the highest composite score favors components that correlate with the composite. Selection can also change relative sub-score patterns, so we treat these patterns as descriptive. Interplanetary shock rows lead the compressive sub-scores, including magnetic and plasma reconstruction errors and pressure jumps, consistent with Rankine-Hugoniot conditions. ICME rows lead the $B_t$ inconsistency, Alfv\'{e}n Mach deviation, and entropy jump scores. These responses are consistent with known ICME properties, including ordered fields and low Alfv\'{e}n Mach numbers. The $\beta$ score responds to changes in beta rather than its level, so shocks can lead that score. The $B_t$ residual also increases with field strength, so the high ICME values partly reflect the ejecta's higher $|\mathbf{B}|$. These differences are qualitative: the plotted detected event sample contains 27 interplanetary shocks, 14 ICMEs, and only 3 high-speed streams. A Kruskal-Wallis test across the three types finds one of the twelve plotted sub-scores significant: the total pressure jump at $p = 0.03$ unadjusted. It fails the Bonferroni threshold of $p < 0.004$ for twelve tests.

We extended this analysis, initially conducted for one year, to PISCES-Opto on the full 2018 to 2024 test set. Of 848 events with scored coverage, 194 have at least one alarm at the 99th percentile threshold. Each detected event contributes the window with its maximum composite score.

An initial test suggested that the sub-scores contained information about event type. Eight of fifteen sub-scores discriminated among the three catalog types at a Kruskal-Wallis $p$ value below $0.05/15$ after Bonferroni correction: the two reconstruction scores, total pressure and plasma beta jumps, magnetic field rotation, the reconstruction and input temperature and velocity scores, and the total pressure residual. This analysis over the full test period includes only the 194 events detected by PISCES-Opto, about one quarter of the 848 events with scored coverage.

Eight of the fifteen sub-scores separate the three catalog event types under that test. That count does not hold up, and we therefore withdraw that count and the inference based on it. Conditioning on an alarm favors ICMEs because 69\% of ICMEs trigger an alarm, compared with 25\% of shocks and 11\% of high-speed streams. The comparison also uses two runs of the same training setup, with each model contributing a different number of alarmed events. Selecting peak windows introduces a further bias, examined below. Without conditioning on alarms, the sample size remains fixed at 848 and the count is 13 or 14 out of 15. The untrained runs in Section~\ref{sec:untrained_control} also give 11 to 14 out of 15 on the same test.

A second issue is the selection criterion itself. Each event contributes sub-scores from the window where its composite score is maximal, but event types offer different numbers of candidate windows: ICME intervals have a median of 1,519 windows, compared with 344 for shocks and 361 for high-speed streams. More candidate windows create greater potential selection bias when a sub-score is associated with the composite. Replacing sub-scores with independent noise and applying the same rule at each event's actual window count shows the effect. The null at correlation zero is $\eta^2_H = 0.000$, and no draw passes Bonferroni correction. At 0.6 association, it is 0.011 and 47\% of draws pass; at 0.8, it is 0.029 and 97\% pass; with an identical sub-score and composite, it reaches 0.123 with every draw passing Bonferroni correction. Reconstruction components have unit weights in the composite, while physics and derived components have weights of a few thousandths. The reconstruction terms therefore dominate the selected score range. We do not use counts based on peak windows as evidence, and we do not treat the withdrawal as uniform: the null shows that selection bias depends on the correlation between each sub-score and the composite. Median aggregation removes the tested selection effect from choosing each maximum window in this null, giving $\eta^2_H = 0.000$ at each correlation. It does not rule out other sources of bias.

After withdrawing the count based on peak windows, we compare component effects under median aggregation. Ten or eleven of fifteen sub-scores remain significant with Bonferroni correction for each trained run, compared with seven to ten for each untrained draw. The ordering of components is more informative than this count. We order the fifteen components by effect size and correlate those rankings across runs. The eleven trained runs have a mean Spearman $\rho$ of 0.88 within their group, the six untrained runs have 0.53, and pairs with one trained and one untrained run have 0.28. Because these pairwise comparisons share runs, we permute trained and untrained status instead of testing pairs independently. The observed difference is the largest among 12{,}376 such permutations, $p = 8\times10^{-5}$, the smallest attainable permutation $p$ value.

Four of the fifteen sub-scores depend entirely on the input window and therefore have the same value for every model under median aggregation. They cannot separate trained from untrained in this analysis. Dropping these shared quantities increases the observed difference between trained and untrained rankings. Across the remaining eleven sub-scores, the correlation is 0.83 for trained runs and $-0.03$ for trained versus untrained runs. Across the nine components that depend on reconstruction, it is 0.80 and $-0.23$, respectively.
Training and physics-informed loss are distinct factors. The three runs from Section~\ref{sec:ablation} trained without physics terms, whose loss drops all seven constraints, have a mean Spearman $\rho$ of 0.937 within their group and correlate with runs trained using physics terms at 0.913. That value is close to the correlation among runs trained using physics terms. Correlations between trained and untrained runs are much smaller, 0.28 for the full trained group and 0.242 for the group without physics terms. These results associate the ordering with trained weights. Physics terms in the objective alone do not explain it. We report the ordering as a property of the trained model and do not claim that the physics-informed loss produced it. In a separate inference comparison, median aggregation gives a correlation of 0.80 between the same checkpoint scored with full skips and with attenuation, close to the correlation between two trained runs. Within this comparison, ordering depends more on the trained model than on inference setup. We report results for the full-skip setup. Effect sizes are small, with the largest at $\eta^2_H \approx 0.15$.

The fixed rule for mapping those sub-scores to event types performs only marginally better than a rule that always chooses the majority class. It labels 80 of the 194 alarmed events correctly, against 79 for a rule that labels every event as an interplanetary shock. It correctly labels 53 of the 79 shocks and none of the 46 high-speed streams. Among ICMEs, the rule assigns the interplanetary shock criterion to 40 of 69, one of the two ICME criteria to 27, and the compression criterion to 2.

We also isolate the contribution of the thirteen physical and derived sub-scores at inference by retaining only the two reconstruction error terms. For the checkpoint trained with all seven physics terms, scoring only the two reconstruction terms gives average precision 0.346 versus 0.365 with the full composite. Event POD is 0.250 versus 0.229.

The comparison of training objectives uses three trained runs per condition in a two by two design, with each run scored using both the full composite and reconstruction error alone. Removing physics terms from the objective while retaining them in the composite gives a mean of 0.257. Removing them from both gives 0.217, against 0.304 for the condition with physics terms in training. Full removal costs 0.087 in average precision, one and a half times the standard deviation among the three runs in that condition. The objective accounts for most of this difference. The two removals are not additive. Changes from using reconstruction error alone for scoring are 0.040 for the condition without physics terms in training and 0.029 for the condition with physics terms in training. For the condition with physics terms in training, score terms are worth 0.029 in average precision. The two training conditions also differ in their observed spread of average precision across runs. The condition without physics terms has a standard deviation of 0.007 against 0.055 for the condition with physics terms in training, although three runs per condition do not support estimating a variance ratio.

\section{Discussion}
\label{sec:discussion}

\subsection*{Contributions of Decomposition}
These comparisons do not establish higher aggregate detection performance for PISCES. PISCES decomposes the anomaly score into fifteen reconstruction, physics, and derived sub-scores. The cited supervised ICME detectors train on binary labels derived from catalog intervals and segment time series \cite{rudisser2022automatic, rudisser2026arcane}. PISCES is trained to reconstruct unlabeled solar wind windows. Its sub-scores quantify reconstruction discrepancies and departures from specified physical relationships.

Sub-score effect size rankings are more consistent within the trained group than within the untrained group, with mean Spearman correlations of 0.88 and 0.53, respectively. This association with training does not identify what the networks learned about ambient wind. The ordering remains similar when physics terms are removed from the objective, so it cannot be assigned to those terms alone. The decomposition does not determine event types. A rule based on the sub-scores is only slightly more accurate than always predicting the majority class. Mapping physical changes to event classes remains an open modeling task.

We treat the contribution as a description of variables that distinguish events from ambient wind. It does not map those variables to event classes. Its practical value for forecasters remains to be tested.

\subsection*{Effects of Training}
Four untrained models shared the same architecture, normalization statistics, and scoring procedure, but had different random weight initializations. With the skip paths set to scale 1.0 at inference, their mean average precision was $0.337 \pm 0.005$, higher than that of ten of the eleven trained runs. Ranking windows by score without setting a threshold does not show that training improves ranking.

For all eleven trained runs, selecting the attenuation scale on validation data improves test average precision by $0.070 \pm 0.028$ on average. In the four untrained runs used for this comparison, average precision changes by only $0.0001 \pm 0.0003$. We designed this comparison after inspecting test set results, so it is retrospective even though attenuation was selected on validation data. The ordering of sub-scores also differs between trained and untrained runs. The three runs trained without physics terms produce almost the same sub-score ordering as runs trained with physics terms.

The untrained comparison does not identify the mechanism, and retraining without skip connections does not resolve the ranking difference. The resulting models omit the parameters that fuse skip features and have fewer parameters. This comparison cannot tell how much of the ranking gain comes from training with skip connections and how much comes from attenuating them during inference. The ablation study of physics loss groups and the comparison of three loss functions across three runs per condition show no consistent increase in average precision from adding physics terms. With each run's threshold set to the 99th percentile of its validation scores, trained runs evaluated with skip scale 1.0 cover more events, although their alarm counts and durations differ. Event coverage is higher for trained runs when normalized by alarm windows and higher for untrained runs when normalized by alarm episodes. Because these two normalizations point in opposite directions, event coverage alone cannot establish a training benefit.

\subsection*{From Detection to Early Warning}
The warning analysis replaces the centered 15-minute median with a trailing 15-minute median, so each output uses the current score and preceding scores only. For seed 42, with attenuation selected on validation data, average precision falls from 0.406 to 0.403. This evaluates filtering of historical OMNI scores. It does not establish end-to-end data availability or live-feed latency.

With full skips, three trained runs have more sudden-commencement matches than four untrained runs in raw counts and when normalized by alarm windows or alarm episodes. The mean match rate per alarm episode across runs is 0.0771 for trained runs and 0.0675 for untrained runs, a 14\% relative increase. Coverage for anchored shocks reverses between normalization by alarm windows and normalization by alarm episodes, so we withdraw it as evidence of a training advantage. For sudden commencements, median estimated leads are 23.9 to 27.9 minutes for trained runs and 11.1 to 12.2 minutes for untrained runs. For anchored shocks, median estimated margins before predicted ground onset are $+17$ to $+18$ minutes for trained runs and $-7$ to $-3$ minutes for untrained runs. These statistics summarize different subsets of matched events across groups and do not establish a paired timing effect.

The raw ACE analysis keeps the checkpoint, score weights, and OMNI normalization statistics fixed. Only the alarm threshold at the 99th percentile is derived from the 2016 to 2017 ACE validation scores after smoothing with a trailing median. Gaps in instrument coverage leave 83 of 112 sudden-commencement events with fewer than 60 scored minutes in the two hours before onset. The main analysis uses OMNI observations from multiple spacecraft. OMNI assigns upstream observations timestamps based on their predicted arrival at the modeled bow shock nose. These timestamps do not show when spacecraft observed the solar wind, when data were received, or when alerts were issued. Prospective validation needs observation, receipt, and alert timestamps.

\subsection*{Limitations}
PISCES is compared with baselines under unequal tuning in Table~\ref{tab:main_results}. We selected PISCES from 27 retained trained configurations after inspecting corrected test metrics. Its inference setting was then selected on validation data. We did not tune the baseline configurations shown in Table~\ref{tab:main_results}. Each uses one fixed configuration, mostly library defaults. We separately selected the Isolation Forest tree count and subsample size on validation data using the same average precision metric and split as our inference setting. This raises its test average precision from 0.358 to 0.373. The observed gain is 0.015 in average precision. Both selected parameters lie at the upper edge of the search grid, but a wider validation search need not improve test performance. The Anomaly Transformer implementation omits the published minimax procedure and ranks by reconstruction error. It is a transformer reconstruction baseline rather than the reference method. The comparison favors PISCES because we tuned it more extensively than those table configurations.

The training and test years also confound solar cycle phase, spacecraft source, and data coverage. OMNI combines time-shifted measurements from different upstream spacecraft. ACE supplies 44\% of the minutes with usable magnetic field data during training but only 9\% during the test period. It stops contributing after 13 March 2021. More than half of the minutes in the test period fall after ACE stops contributing, leaving Wind as the only source. ACE's share also varies within the training period, from 70\% ACE in 2005 to 18\% in 2011. Coverage of all seven features falls from 75\% during training to 69\% during testing, with 59\% in 2023 and 60\% in 2024. The full-skip comparison does not establish a ranking benefit from training. The exact permutation test comparing eleven trained runs with full skips to four untrained runs gives only marginal evidence for a difference in average precision ($p = 0.054$). Only the homogeneous subset of eight runs reaches $p = 0.008$, but we cannot claim that training improves ranking when scores are ranked without a threshold and skips are at full strength. OMNI lacks electron temperature, so the pressure estimate excludes electron thermal pressure and can vary in bias across wind regimes \cite{newbury1998electron,schrijver2009plasma}. The entropy term separately penalizes temporal changes in the reconstructed proton entropy proxy. It does not follow the same solar wind material as it moves or compare changes in the pressure estimate with changes in the input. Directional discontinuities and turbulence may cause false alarms, but this mechanism was not tested directly \cite{tsurutani1979interplanetary,soding2001radial,verscharen2019multiscale}. Inference in real time uses L1 measurements, while OMNI shifts upstream measurements to their predicted arrival at the modeled bow shock nose \cite{papitashvili2020omni}. The ACE raw data analysis keeps the model and score weights fixed and sets only its 99th percentile threshold from ACE validation scores. Event coverage is incomplete, and this analysis does not simulate delays before an operational system receives data or the order in which data are received.

For PISCES-Mono, average precision depends on the matching tolerance for point events. For PISCES-Opto, the fixed rule that maps sub-scores to event types only slightly exceeds the majority class rule, as reported in Section~\ref{sec:decomposition}.

\subsection*{Future Work}
Models using input windows of 30, 60, and 120 minutes could test whether shorter windows retain shock signatures and longer windows provide more context around ICME boundaries and stream interaction regions \cite{cash2014characterizing,cane2003interplanetary,grandin2019properties}. The window length sets the amount of input. Solar wind events may still last hours or days.

Across three runs, training with an objective based on derived quantities differed little in average precision from training with reconstruction error alone. Synchronized electron and proton measurements would allow the total pressure and plasma beta estimates to include electron thermal pressure. An Alfv\'{e}nicity measure could also account for magnetic field polarity. That measure needs synchronized vector velocity, magnetic field, mass density, and a chosen time interval. Solar wind fluctuations over large scales are often Alfv\'{e}nic, especially in fast wind \cite{belcher1971large,verscharen2019multiscale}. It may not separate benign discontinuities. Each experiment should set random seeds in advance and compare PISCES with models of the same size trained only on reconstruction error and with untrained networks.

Operational testing needs prospective records of source spacecraft, observation time, receipt time, and alert time for live Deep Space Climate Observatory (DSCOVR) and Advanced Composition Explorer (ACE) data. Data should be processed in the order received. Blind testing needs a period unused during development and an analysis procedure fixed beforehand. A randomized forecaster study should compare the fifteen sub-scores with one score from reconstruction error on the same unlabeled cases.

\section{Conclusion}
\label{sec:conclusion}

We presented PISCES, a convolutional autoencoder trained on eleven years of unlabeled OMNI data to model solar wind and flag anomalous windows. Its score has fifteen sub-scores for reconstruction, physics, and derived quantities. PISCES-Opto reaches average precision of 0.365 across pooled score windows from 2018 to 2024, with scored coverage overlapping 848 catalog events. A bootstrap over calendar months gives a 95\% confidence interval of 0.306 to 0.419. We selected this checkpoint from 27 retained trained configurations after inspecting corrected test metrics. It is corrected retrospective evidence, not blind test evidence. Selecting skip attenuation on validation raises average precision to 0.406. Across eleven runs, the mean rises from $0.306 \pm 0.029$ to $0.376 \pm 0.020$. The attenuation analysis was designed after test inspection.

With full skips, four untrained runs reach $0.337 \pm 0.005$, exceeding ten of eleven trained runs. This comparison does not establish a ranking benefit from training. In two untrained draws evaluated on 7,581 finite 2018 windows advanced 60 minutes at a time, decoder output energy is below 0.5\% of input energy. Reconstruction error then approaches the mean squared standardized input, so untrained weights can still rank windows. Attenuation increases average precision by $0.070 \pm 0.028$ for trained runs and $0.0001 \pm 0.0003$ for untrained runs. This difference does not isolate an effect from the physics terms in training.

PISCES decomposes its anomaly score into reconstruction, physics, and derived quantities. These results do not establish aggregate superiority. Across the three warning analysis runs, after smoothing with a trailing median and separate 99th percentile thresholds, attenuation pairs an alarm with 20 to 22 of 112 sudden commencements, compared with 32 to 39 under full skips. Median estimated leads are 18 to 19 minutes with attenuation and 24 to 28 minutes with full skips. For this comparison, each median describes that configuration's own subset of matched events. The numbers of alarms differ, so the comparison does not establish a paired attenuation effect. Selecting the window with the highest composite score can also bias event type comparisons because event types have different numbers of candidate windows. Median aggregation removes the tested maximum selection effect, but cannot remove every possible source of bias.

Future work should test different window lengths and evaluate Alfv\'{e}nicity using synchronized vector measurements. Operational evaluation requires live DSCOVR and ACE data processed in receipt order, with observation, receipt, and alert times recorded. Blind evaluation needs a period unused during development and an analysis procedure fixed beforehand. A randomized forecaster study should compare fifteen sub-scores with a single score based on reconstruction error on the same unlabeled cases.

\newpage

\begin{ack}
We thank the NASA Space Physics Data Facility (SPDF) for providing the high-resolution OMNI data used here, and the Wind and ACE mission teams for the upstream solar wind measurements underlying OMNI. We also thank the NASA Community Coordinated Modeling Center (CCMC) for the DONKI event catalogs and the NOAA Space Weather Prediction Center (SWPC) for its real-time solar wind data products. We thank Ian G. Richardson and Hilary V. Cane for their near-Earth ICME catalog, publicly available through the ACE Science Center at Caltech. Our sudden-commencement validation uses event lists published by Observatori de l'Ebre in Spain, the collaborating institute of the International Service of Geomagnetic Indices (ISGI) for rapid magnetic variations from magnetic observatory data. We also thank the International Real-time Magnetic Observatory Network (INTERMAGNET) and ISGI.
\end{ack}

\bibliographystyle{ieeetr}
\bibliography{references}

\newpage
\appendix

\section{Data Split}
\label{app:data}

\begin{table}[H]
\caption{Summary of the data split. Window counts exclude windows that still contain missing values after preprocessing. Training and early stopping validation advance the 60-minute window by 15 minutes. Test evaluation and validation threshold estimation advance it by one minute.}
  \label{tab:data_summary}
  \centering
  \vspace{6pt}
  \resizebox{\textwidth}{!}{%
  \begin{tabular}{lllrr}
    \toprule
    Set & Period & Purpose & Records & Windows \\
    \midrule
    Training & 2005 to 2015 & Fit distribution of unlabeled historical data & $\sim$5.8M & $\sim$331K \\
    Validation & 2016 to 2017 & Calibrate alarm threshold and guide early stopping & $\sim$1.1M & 56,565 \\
    Test & 2018 to 2024 & Evaluate from cycle 24 minimum to cycle 25 rise & $\sim$3.7M & 2,969,144 \\
    \bottomrule
  \end{tabular}%
  }
\end{table}

\vspace{1em}

\section{Physics Constants}
\label{app:constants}

\begin{table}[H]
\caption{The physical constants used by the physics losses are expressed in OMNI units (nT, cm$^{-3}$, K, km/s, nPa). In SI units, $k_B = 1.38065 \times 10^{-23}$ J/K and $\mu_0 = 4\pi \times 10^{-7}$ H/m. The factor for Alfv\'{e}n speed is the only listed constant not used in the training loss. It enters only the Alfv\'{e}n Mach score.}
  \label{tab:constants}
  \centering
  \vspace{6pt}
  \begin{tabular}{llll}
    \toprule
    Constant & Symbol & Value & Unit \\
    \midrule
    Boltzmann constant (thermal pressure)
      & $k_B$ & $1.3807 \times 10^{-8}$ & nPa$\cdot$cm$^3$/K \\
    Reciprocal of $2\mu_0$ (magnetic pressure)
      & $1/(2\mu_0)$ & $3.9789 \times 10^{-4}$ & nPa/nT$^2$ \\
    Dynamic pressure factor ($\frac{1}{2}m_p n_p V^2$)
      & n/a & $8.363 \times 10^{-7}$ & nPa/(cm$^{-3}\cdot$(km/s)$^2$) \\
    Alfv\'{e}n speed factor
      & n/a & $21.81$ & km/s per nT/$\sqrt{\text{cm}^{-3}}$ \\
    Parker spiral $\Omega_\odot r$
      & $\Omega_\odot r$ & 428.7 & km/s \\
    Slope of $T$ versus $V$ fit (Elliott et al.)
      & n/a & 486.5 & K per km/s \\
    Intercept of $T$ versus $V$ fit
      & n/a & $-124{,}760$ & K \\
    Adiabatic index (proton gas)
      & $\gamma$ & $5/3$ & dimensionless \\
    Entropy exponent
      & $\gamma - 1$ & $2/3$ & dimensionless \\
    \bottomrule
  \end{tabular}
\end{table}

\newpage

\section{Hyperparameters}
\label{app:hyperparams}

\begin{table}[H]
\caption{Architecture, training, and inference parameters for PISCES-Opto. The seven loss coefficients, $pw_{\max}$, warmup and ramp periods, encoder filter settings, decoder kernel sizes, skip dropout, learning rate, patience, and parameter count come from the saved checkpoint and its training configuration. The remaining values come from the training scripts. The squeeze-and-excitation (SE) reduction ratio is 4. This gives hidden widths of 8, 4, and 4 for channel counts 32, 16, and 8.}
  \label{tab:hyperparams}
  \centering
  \vspace{6pt}
  \begin{tabular}{ll}
    \toprule
    Parameter & Value \\
    \midrule
    Window size & 60 minutes \\
    Training step & 15 minutes \\
    Inference step & 1 minute \\
    Encoder filters & [32, 16, 8] \\
    Max pool factors & [2, 2, 3] \\
    Encoder kernel sizes & [7, 5, 3] (same padding) \\
    Decoder kernel sizes & [3, 5, 7], reversed relative to encoder \\
    Dilations & [1, 2, 1] \\
    Bottleneck attention & Squeeze-and-excitation (reduction ratio 4) \\
    Encoder attention & Squeeze-and-excitation at each level (reduction ratio 4) \\
    Decoder attention & Squeeze-and-excitation at each level (reduction ratio 4) \\
    Skip connection dropout & 0.8 during training only \\
    Bottleneck shape & [8, 5] \\
    Compression ratio & 10.5:1 \\
    Total parameters & 12,555 \\
    Batch size & 256 \\
    Learning rate & $10^{-3}$ \\
    Optimizer & Adam \\
    Weight decay & $10^{-4}$ \\
    Gradient clipping & max norm 1.0 \\
    Learning rate schedule & Cosine annealing after warmup \\
    Minimum learning rate & $10^{-6}$ \\
    Early stopping patience & 25 epochs \\
    $\lambda_{\text{Bt}}$ & $5 \times 10^{-4}$ \\
    $\lambda_{\text{ent}}$ & 0.5 \\
    $\lambda_{\text{pres}}$ & 0.3 \\
    $\lambda_{\text{tv}}$ & 0.003 \\
    $\lambda_{\text{parker}}$ & $3 \times 10^{-4}$ \\
    $\lambda_{\beta}$ & 0.03 \\
    $\lambda_{\text{dyn}}$ & 0.05 \\
    Physics warmup & 10 epochs using MSE only \\
    Physics ramp & 10 epochs in ten equal steps from $0.1\,pw_{\max}$ to $pw_{\max} = 0.30$ \\
    Offline score smoothing & 15-minute centered median \\
    Threshold percentile & 99th \\
    Maximum interpolation gap & 15 minutes \\
    Forward fill limit & 2 minutes \\
    Random seed & 42 \\
    \bottomrule
  \end{tabular}
\end{table}

\newpage

\section{Sensitivity to Matching Tolerance}
\label{app:match_window}

We use a 3-hour match tolerance in the evaluation criterion (Section~\ref{sec:experiments}). Table~\ref{tab:match_window} shows how PISCES-Mono metrics vary with tolerance on the 2018 to 2024 test set. Changing the matching tolerance changes event labels and hit assignments while leaving detection timestamps fixed. A wider window counts more detections as hits. Detected events increase from 61 to 201 of 848, PR-AUC rises from 0.220 to 0.419, and the false alarm ratio falls from 0.557 to 0.188. Window POD decreases from 0.079 to 0.045 because the number of positively labeled windows grows faster than the number of true-positive alarm windows. At 3 hours, PR-AUC is 0.322, matching PISCES-Mono in Table~\ref{tab:main_results}.

\begin{table}[H]
\caption{Sensitivity to match window tolerance for PISCES-Mono on the 2018 to 2024 test set. Precision-recall area under the curve (PR-AUC) is independent of the alarm threshold but depends on the event labels defined by matching tolerance. Probability of detection (POD) and false alarm ratio (FAR) use the 99th percentile validation threshold. ``Detected'' gives the number of catalog events with at least one detection in the matching window out of the 848 events eligible for coverage.}
  \label{tab:match_window}
  \centering\small
  \vspace{6pt}
  \begin{tabular}{lccccc}
    \toprule
    Match window & PR-AUC & POD & FAR & Positive window rate & Detected \\
    \midrule
    $\pm 0.5$\,h & 0.220 & 0.079 & 0.557 & 5.7\% & 61/848 \\
    $\pm 1$\,h & 0.234 & 0.070 & 0.521 & 7.0\% & 82/848 \\
    $\pm 2$\,h & 0.281 & 0.062 & 0.422 & 9.5\% & 130/848 \\
    $\pm 3$\,h & 0.322 & 0.056 & 0.352 & 11.9\% & 155/848 \\
    $\pm 6$\,h & 0.419 & 0.045 & 0.188 & 18.3\% & 201/848 \\
    \bottomrule
  \end{tabular}
\end{table}

\end{document}